\newif\ifcolm
\colmtrue

\PassOptionsToPackage{hyphens}{url} %

\ifcolm
  \documentclass{article}
  \usepackage[preprint]{colm2026_conference} %
  \usepackage{hyperref} %
  \usepackage{afterpage} %
\else
  \documentclass[letterpaper]{article} %
  \usepackage{aaai2027}  %
\fi
\usepackage{url} %
\usepackage{graphicx} %
\usepackage{natbib}  %
\usepackage{caption} %
\usepackage{booktabs}

\usepackage{latexsym}
\usepackage{microtype}
\usepackage{inconsolata}
\usepackage{subcaption}
\usepackage{booktabs}
\usepackage{amsmath}
\usepackage{amssymb}
\usepackage{amsthm}
\newtheorem{definition}{Definition}
\usepackage{xcolor}
\usepackage{multirow}
\usepackage{fontawesome5} %
\usepackage[capitalize,noabbrev]{cleveref}

\ifcolm\else
\fi

\title{Latent Programming Horizons in Coding Agents}

\ifcolm
\author{%
    Andr\'e Silva, Han Tu, Martin Monperrus \\
    KTH Royal Institute of Technology \\
    Stockholm, Sweden \\
    \texttt{\{andreans, htu, monperrus\}@kth.se}
}
\else
\author{
    André Silva,
    Han Tu,
    Martin Monperrus
}
\affiliations{
    KTH Royal Institute of Technology\\
    Stockholm, Sweden\\
    \texttt{\{andreans, htu, monperrus\}@kth.se}
}
\fi

\newcommand{\CodeURL}{https://github.com/ASSERT-KTH/program-probes}
\newcommand{\DataURL}{https://huggingface.co/datasets/ASSERT-KTH/latent-programming-horizons-trajs}
\newcommand{\hflogo}{\raisebox{-0.2ex}{\includegraphics[height=1.0em]{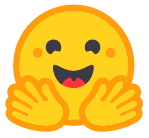}}}
\ifcolm
  \newcommand{\CodeLink}{\href{\CodeURL}{\faGithub\,\textbf{Code}}}
  \newcommand{\DataLink}{\href{\DataURL}{\hflogo\,\textbf{Data}}}
\else
  \newcommand{\CodeLink}{\faGithub\,\textbf{Code}: \url{\CodeURL}}
  \newcommand{\DataLink}{\hflogo\,\textbf{Data}: \url{\DataURL}}
\fi

\definecolor{propSyn}{HTML}{E69F00}%
\definecolor{propSem}{HTML}{009E73}%
\definecolor{propRed}{HTML}{CC79A7}%
\definecolor{propReg}{HTML}{8C6D31}%
\newcommand{\symSyn}{\textcolor{propSyn}{\ensuremath{\blacktriangle}}}
\newcommand{\symSem}{\textcolor{propSem}{\ensuremath{\bullet}}}
\newcommand{\symRed}{\textcolor{propRed}{\ensuremath{\blacksquare}}}
\newcommand{\symReg}{\textcolor{propReg}{\ensuremath{\blacklozenge}}}
\newcommand{\Syntactic}{\symSyn\,\textcolor{propSyn}{\textsc{Well-formedness}}}
\newcommand{\Semantic}{\symSem\,\textcolor{propSem}{\textsc{Full Correctness}}}
\newcommand{\RedFail}{\symRed\,\textcolor{propRed}{\textsc{Partial Correctness}}}
\newcommand{\Regressions}{\symReg\,\textcolor{propReg}{\textsc{Regression}}}

\newcommand{\scfigwidth}{\ifcolm 0.62\linewidth\else\columnwidth\fi}

\begin{document}

\maketitle

\begin{abstract}
A coding agent solving a software-engineering task spends dozens of steps reasoning, editing code, and running tests, yet little is known about what the underlying language model internally represents about the program it is working on.
We show that the residual streams of language models under coding agents linearly encode properties of the evolving program: a logistic-regression probe on hidden states is able to decode whether the current code parses, passes its test suite, reduces the number of failing tests, and introduces regressions, reaching AUC up to $0.83$ for correctness across two models and two benchmarks.
Our second finding is more surprising: these representations run ahead of the agent's own edits.
Probes trained to predict the outcome of future edits (before they are materialized and written on disk) achieve performance above chance up to roughly 25 steps in advance.
We call this the agent's \emph{latent programming horizon}.
As a proof of external validity, we show that the probes transfer across benchmarks without retraining.
Our positive results open calls for more research in mechanistic interpretability of coding agents.
\end{abstract}

\ifcolm
\begin{center}
\CodeLink \quad\textbar\quad \DataLink
\end{center}
\else
\fi

\newcommand{\figLatent}{%
\begin{figure*}[t]
  \centering
  \includegraphics[width=\textwidth]{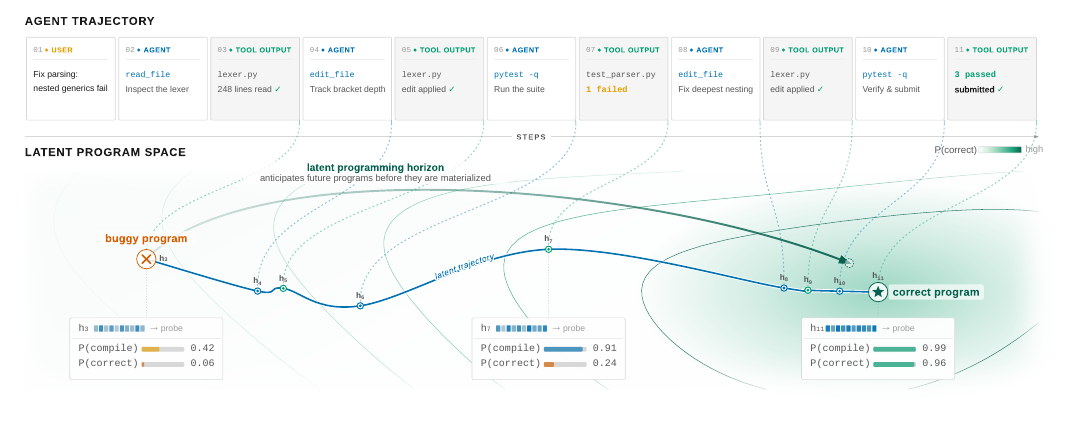}
  \caption{
    A coding agent iteratively edits code and runs tests (top); the hidden states at each step trace a corresponding path through the latent program space (bottom). The space is shaded by probe-estimated $P(\text{correct})$, with a warm basin marking the final region where the final program is fully correct and the coding task is successful.
    Our experiments show that linear probes trained on hidden states decode program properties such as full correctness well above chance throughout the agentic trajectories.
    Beyond current state, the latent program representation captures future programs' properties: this demonstrates that coding agents have a \emph{latent programming horizon}.
  }
  \label{fig:latent}
\end{figure*}%
}
\ifcolm
  \afterpage{\figLatent}
\else
  \figLatent
\fi

\section{Introduction}
\label{sec:intro}

Coding agents are increasingly used to solve complex software engineering tasks.
Over many steps, they read files, reason about the problem, edit code, run tests, and revise their changes, iterating until the task is complete.
Despite intense interest in building such agents, strikingly little is known about how the underlying language model represents the code it is working on.
In this paper, we study the following question:
\emph{what does a coding agent internally represent about the program it is editing?}

Prior work in other domains gives reason to expect a rich answer.
Transformers trained on board games develop linearly decodable internal representations of the game state \citep{liemergent, nanda2023emergent, karvonenemergent}.
Models trained on synthetic grid-world programs encode both the current and future semantic state of those programs \citep{jin2024emergent}.

In the software domain, prior work \citep{spiess2025internal, correctness2025internal, autoprobe2025} has established that the correctness of a single generated function is linearly decodable from a model's hidden states.
Yet, these papers assume single-step generation where the full program sits in context and never changes.
None of them is about coding agents iteratively editing an only partially observed real codebase, over dozens of steps, where the model's hidden state shifts with every edit. This is the problem we address in this paper.

To study this arguably difficult question, we proceed as follows.
We systematically collect agentic trajectories, extract the residual stream of the model at each agentic step, and train linear probes to decode properties of the program being edited.
We consider two open-weight models running \texttt{mini-swe-agent}~\citep{yang2024sweagent} on two benchmarks (\texttt{SWE-Bench-Verified}~\citep{jimenez2024swe} and \texttt{SWE-Bench-Pro}~\citep{deng2025swe}).
Two main findings emerge.

\paragraph{Coding agents encode the current program.}
Linear probes recover hard program properties. We show that the probes capture whether the current program parses, passes its test suite, reduces the number of failing tests, and introduces regressions, all well above a sound shuffled-label control.
The signal persists throughout the trajectory and is strongest in the model's intermediate layers.

\paragraph{Coding agents anticipate future programs.}
Beyond tracking the present, the agents represent programs that have not yet materialized.
Probes trained to predict the outcome of a future edit succeed up to roughly 25 steps in advance.
This is the coding analogue of learned look-ahead in game-playing networks \citep{jenner2024chess,taufeeque2024sokoban}.
Coding agents have an idea of the shape of the future programs they will write, well before those changes are materialized with edits.

\paragraph{}
To the best of our knowledge, we are the first to show that coding agents maintain and continually update latent representations of programs and their semantic properties (\Cref{fig:latent}). Furthermore, we are the first to report on a long-term programming horizon in the latent space, with strong evidence from ahead-of-chance prediction capabilities. 

\paragraph{Contributions.}
\begin{itemize}
  \item We propose a novel experimental protocol to study how coding agents maintain a representation of programs in the latent space.
  \item We show that linear probes on the hidden states of coding agents are able to decode four hard properties of programs. Our experimental results are based on real-world codebases and multi-step trajectories (median = 52 steps). We demonstrate that these probes are calibrated and transfer across benchmarks without retraining.
  \item We show that the latent space captures the programming horizon. The latent program representation grounds future edit prediction: probes can predict the outcome of an edit $k$ steps ahead, and this holds for up to $k = 25$ steps.
\end{itemize}

\ifcolm
\newpage
\else
\fi

\section{Latent Program Representations}
\label{sec:latentrepr}
Consider a coding agent solving a programming task by iteratively searching and reasoning about the codebase, making edits, running tests, and using the feedback to guide its next steps.
Beyond the tokens the underlying model emits and observes, such an agent may encode a latent program representation, an internal account of the program that it updates and revises. This representation, invisible to the harness and the user at the token level, is the object we study in this paper.

This representation, if it exists, lives somewhere in the internal calculation of the model.
Indeed, transformer models maintain a running hidden-state vector, called the residual stream, which each layer reads from and writes an additive update back to, carrying information from the input through to the output. 
We write $\mathbf{h}_t^\ell \in \mathbb{R}^d$ for the residual stream at token position $t$ after layer $\ell$. Because every layer communicates through it, the residual stream is the standard target for probing what a model internally represents
\citep{elhage2021mathematical}.
In this paper, we study whether the residual stream contains latent program representations.

\begin{definition}[Latent program representation]
A latent program representation is what is encoded about a program in the residual stream $\mathbf{h}_t^\ell$. It is captured independently from the surface tokens the agent has emitted or will emit.
\end{definition}

The residual stream does not only contain the latent program representation. It also carries information about the task at hand and the past conversation, so the latent program representation is tangled with other signals in the hidden state, and recovering it is the job of the probes we introduce later, see \autoref{sec:probing}.

\begin{definition}[Latent program space]
The latent program space is the manifold of the hidden-state space $\mathbb{R}^d$ along which latent program representations are encoded. As the agent edits code and runs tests, the hidden state moves along this manifold, forming a latent program representation trajectory.
\end{definition}

At each step, the hidden state is the point on this manifold, it captures the current program. A capable agent, however, might not only reason about the present. If the model is planning ahead, its current hidden state should also encode properties of the future programs, the ones it might write several steps ahead. We call the reach of this planning in the latent space the ``latent programming horizon''.

\begin{definition}[Latent programming horizon]
The latent programming horizon is the extent to which future edits are already encoded in the current residual stream. 
We define the latent programming horizon as the capability of predicting program properties of the program at $t$ + $k$ steps in advance, only based on $\mathbf{h}_t^\ell$. We refer to it as the programming horizon at $k$.
\end{definition}

The horizon $k$ captures the depth of future planning. Any prediction about the program at $k=0$ means that the properties of the current program are represented in the latent space
For small $k$ (a short-term horizon), prediction success means the current hidden state already carries properties of edits that are a few steps away, for example, a sequence of edits in the same file.
For large $k$ (a long-term horizon), prediction success means the agent anticipates programs that it will write only much later in the trajectory.

While language models are not explicitly trained to encode program representations in their hidden states, several training signals may push them in that direction.
Next-token prediction implicitly rewards learning program syntax and semantics, as predicting the next line of a function is easier if those are understood.
Beyond next-token training objectives, modern code models are also post-trained with signals based on execution traces \citep{copet2025cwm} or feedback from compilers and test suites \citep{ye2022selfapr}, particularly during reinforcement learning \citep{wei2026swe}. Basically, both pre-training and post-training should contribute to having semantically rich latent representations of the program a coding agent is working on. 

In this paper, we are the first to study whether latent program representations have emerged in the latent space of coding models and how coding agents maintain them during software engineering tasks.

\section{Methods}
\label{sec:methods}

\begin{figure}[t]
  \centering
  \includegraphics[width=\scfigwidth]{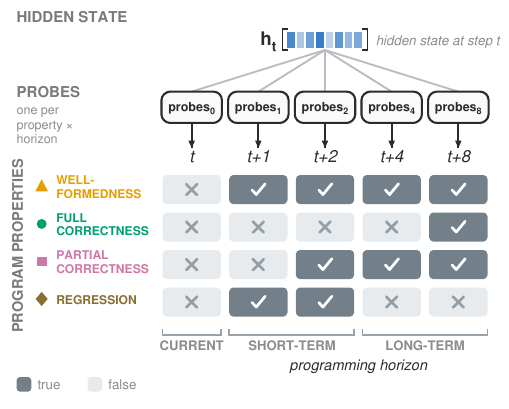}
  \caption{
    Our method to train probes to predict program properties.
    A residual-stream vector $\mathbf{h}_t$, taken at step $t$ of an agentic editing trajectory, is processed by linear probes, parameterized with a lookahead programming horizon $k$.
    Each probe predicts whether the program $k$ steps ahead one of four binary properties (\Syntactic{}, \Semantic{}, \RedFail{}, \Regressions{}); ground-truth labels are computed at each edit event by checking out that program version and evaluating it.
    At $k{=}0$ the probe evaluates whether the latent space maps the program as it currently stands, while probes at larger $k$ measure how far ahead $\mathbf{h}_t$ anticipates the program the agent will eventually write: its latent programming horizon.
    Filled cells mark a property observed in the program (true), light cells an unsatisfied one (false); all computed symbolically from the training trajectories. At inference time, the probes predict future properties, and we measure the prediction accuracy.
  }
  \label{fig:workflow}
\end{figure}

We study what coding agents do during multi-step agentic editing of real codebases, in the latent space.

\subsection{Overview}
We need a way to capture program representations in the hidden state.
For this, we follow the probing literature \citep{alain2017understanding, zhangreasoning} and train linear probes to classify program properties from hidden states.
\Cref{fig:workflow} shows an overview of our setup.
If such probes predict program properties above chance, it shows the model actually encodes those properties in its residual stream.

\subsection{Formal Setup}

Let $\mathcal{M}$ be a language model with $L$ transformer layers.
An agentic trajectory $\mathcal{T}$ for a coding task consists of $S$
interaction steps.
At each step $s$, the agent generates a response $A_s$ given the accumulated context; the tool then produces an output $O_s$.

The full trajectory is the linearised concatenation:
\[
  \mathcal{T} = C_1 \oplus A_1 \oplus O_1 \oplus \cdots \oplus A_S \oplus O_S
\]
where $C_1$ is the initial context (typically system and user prompts), and $\oplus$ denotes token-sequence concatenation.

For each position $t$ in the generated portions $\bigcup_{s=1}^{S} A_s$, let $\mathbf{h}_t^\ell \in \mathbb{R}^d$ denote the residual stream after layer $\ell$.
We index layers from $1$, so that $\mathbf{h}_t^1$ is the residual stream after the first transformer block and $\mathbf{h}_t^L$ after the final block.

An \emph{edit event} occurs at step $s$ when $A_s$ includes a tool call that modifies the codebase.
Let $e_1 < e_2 < \cdots$ denote the token indices of successive edit
events within $\mathcal{T}$.
The edits capture the trial-and-error process of the model to perform the task at hand.

In this paper, a \emph{program property} $\phi$ is a function that assigns to each edit event $e_i$ a binary label $y_i^\phi \in \{0,1\}$ reflecting a property of the resulting program version.
For example, the property `compilability' captures whether a program at a given edit compiles, see \Cref{sec:program-properties} for the list of properties considered in this paper.

Labels are computed externally only at each event, and then kept until the next edit:
every generated token position $t$ with $e_i \le t < e_{i+1}$ receives the label of edit $e_i$.

We also evaluate $\phi$ on the initial program version and treat it as a phantom edit $e_0$ preceding the trajectory, so every position $t$ has a well-defined label even before the first agent edit.

It is convenient to index labels by step rather than by token: let $\sigma(t)$ be the step in which token $t$ is generated (so $t \in A_{\sigma(t)}$), and let $y_s^\phi$ denote the label in effect at step $s$, that is, the label of the most recent edit at or before step $s$. The token-level label is then $y_t^\phi = y_{\sigma(t)}^\phi$.

\subsection{Program Properties}
\label{sec:program-properties}

We define four program properties to study through probing.
We will train probes to predict both current and future values of these properties from the hidden states of the model while working on a program.

\paragraph{\Syntactic{}}
Well-formedness is a necessary precondition for code to execute.
We operationally define it by parsing or compiling the program at a given version.
The label is $1$ if it succeeds and $0$ otherwise.

\paragraph{\Semantic{}}
Full correctness tracks the functional behavior of the program, which is the ultimate goal of code generation and repair.
The label is $1$ if the program passes the benchmark's test suite oracle, and $0$ otherwise.

\paragraph{\RedFail{}}
Rather than asking whether the code is fully correct, this probe captures partial correctness, whether the program is moving in the right direction.
The label is $1$ if the number of failing tests is strictly less than at the beginning of the task, and $0$ otherwise.

\paragraph{\Regressions{}}
A capable agent must attend not only to the tests it is trying to fix but also to those it must not break.
This probe tests whether hidden states carry a signal about side effects of an edit.
The label is $1$ if any test that passed at the beginning of the task now fails, and $0$ otherwise.

\paragraph{}
Together, these four properties span what a coding agent must track to resolve a task: the well-formedness of an edit (\Syntactic{}), its functional correctness (\Semantic{}), its directional progress per the test suite outcomes (\RedFail{}), and its collateral effect on the rest of the repository (\Regressions{}).

\subsection{Probing Program Poperties}
\label{sec:probing}

For each layer $\ell \in \{1, 11, 21, 31, 40\}$, we collect all hidden states $\mathbf{h}_t^\ell$ together with their labels $y_t^\phi$.
To test whether hidden states at layer $\ell$ carry information about $\phi$, we train a logistic regression classifier, $\mathbf{w} \in \mathbb{R}^d$, to predict $y_t^\phi$ from the residual stream $\mathbf{h}_t^\ell$.
We split trajectories, for training and testing, based on task identifier, ensuring no trajectories of the same task appear in more than one split.
As a sanity check, to verify that the probes decode a signal already present in the hidden states rather than fit it themselves \citep{hewitt2019designing}, we also retrain each probe on randomly permuted labels.

The classifiers trained on one benchmark are further used to probe the hidden states of another to check if the signal they extract transfers across benchmarks.
Probe hyperparameters are selected via random search sweeps based on mean validation AUC.

\subsection{Programming Horizon}
\label{sec:lookahead}
To study whether coding agents have a programming horizon, we study whether one can predict the outcome of a future edit before that particular edit appears in their output.
Concretely, we train probes to predict program properties at varying lookahead horizons.

For each property $\phi$ and horizon $k \ge 0$ (in steps), we train a probe on the hidden state $\mathbf{h}_t^\ell$ to predict $y_{\sigma(t)+k}^\phi$, the label of the program $k$ steps later.
We fit a separate probe at each $k$.
We sweep $k$ from $0$ (equivalent to the current-state probe) to $k_{\max} = 50$, excluding the final $k_{\max}$ steps of each trajectory to hold the token population constant across horizons.
If the probe discriminative performance is higher than chance for a given $k$, it means that the coding agent has a programming horizon of length $k$.

\begin{table}[t]
\centering
\footnotesize
\caption{Dataset statistics per model and benchmark. \#Traj.\ counts all agent runs; $\geq$50 counts trajectories reaching that step threshold (the length filter used in the lookahead experiments). Med.\ steps is the median number of steps per trajectory. \#Edits is the total number of code edits across all trajectories. $\geq$2 edits counts trajectories with at least two edits. Med.\ edits is the median number of edits per trajectory. $\#h_t$ is the total number of collected hidden-state vectors. The full trajectory dataset is publicly available: \DataLink.}
\label{tab:dataset_stats}
\resizebox{\columnwidth}{!}{
\begin{tabular}{llrrrrrrr}
\toprule
Model & Benchmark & \#Traj. & $\geq$50 steps & Med.\ steps & \#Edits & $\geq$2 edits & Med.\ edits & $\#h_t$ \\
\midrule
\multirow{2}{*}{\texttt{Laguna}} & Verified & 4,991 & 2,837 & 55 & 14,545 & 2,974 & 2 & 7.5M \\
 & Pro & 6,921 & 4,159 & 56 & 36,648 & 5,273 & 4 & 7.6M \\
\addlinespace
\multirow{2}{*}{\texttt{Qwen3.6}} & Verified & 4,998 & 2,001 & 39 & 9,730 & 2,161 & 1 & 3.8M \\
 & Pro & 5,804 & 3,186 & 54 & 18,557 & 3,682 & 2 & 3.5M \\
\midrule
\multicolumn{2}{l}{Total} & 22,714 & 12,183 & 52 & 79,480 & 14,090 & 2 & 22.4M \\
\bottomrule
\end{tabular}
}
\end{table}

\subsection{Dataset}

We consider two coding agents on two benchmarks.
For agents, we run \texttt{mini-swe-agent} v2.2.8~\citep{yang2024sweagent} with two state-of-the-art medium-sized models, \texttt{Qwen3.6-35B-A3B}~\citep{qwen36plus} and \texttt{Laguna-XS.2}~\citep{abadji2026laguna}.
Both models have a hidden-state dimension of $d = 2048$.
Regarding benchmarks, we use \texttt{SWE-Bench-Verified}~\citep{jimenez2024swe} (all 500 tasks) and \texttt{SWE-Bench-Pro}~\citep{deng2025swe} (all 731 tasks).

For each task, we attempt to generate up to $n = 10$ trajectories.
Agent hyperparameters, including limits, are listed in \Cref{tab:agent_hparams}.
For each trajectory, we record the full program version after every tool call that modifies the codebase, and the results of evaluating the considered program properties.

For each trajectory, we record the hidden states every 5 tokens, for the sake of space.
\Cref{tab:dataset_stats} summarises the resulting dataset; trajectory length distributions (a median of 36k tokens per trajectory) and label prevalence per probe are shown in the Appendix (\Cref{fig:dataset_stats}).
\Cref{tab:dataset_stats} clearly shows that our trajectories are multistep, with a median number of two and a mean of 3.5 edits per trajectory.

\section{Results}
\label{sec:results}

\subsection{Coding agents encode the current program}
\label{sec:results:current}

\begin{figure}[tb]
  \centering
  \includegraphics[width=\scfigwidth]{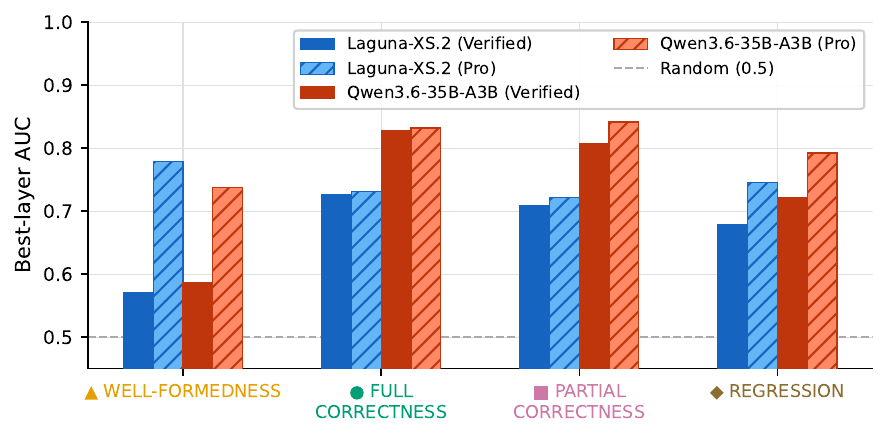}
  \caption{Best-layer AUC for each probe, model, and dataset. Each bar shows the peak AUC across layers. Probes trained on randomly shuffled labels collapse to $0.50$, confirming that the signal is in the representation rather than an artefact of probe capacity. Full per-layer results are in \cref{tab:auc_roc} in the appendix.}
  \label{fig:auc_barplot}
\end{figure}

\Cref{fig:auc_barplot} reports our first main finding.
Every probe, on both models, is able to decode its corresponding program property well above chance.
For example, considering \Syntactic{}, the best-trained probe is for \texttt{Laguna-XS.2}, with an AUC of $0.78$ well above the random baseline of $0.5$.
In other words, the hidden state of the model encodes the program properties of the program it is working on.

\paragraph{Predicting semantic properties is generally stronger.}
The signal is strongest for \Semantic{} (AUC up to $0.83$) and \RedFail{} (AUC up to $0.84$) for \texttt{Qwen3.6-35B-A3B}.
Similar results can be observed for \texttt{Laguna-XS.2}, where \Semantic{} is the strongest on \texttt{SWE-Bench-Verified} (AUC up to $0.73$) and where \Regressions{} reaches an AUC up to $0.75$ on \texttt{SWE-Bench-Pro}.
This means that the model's hidden states capture the semantic changes implied by every edit.
This holds consistently across both benchmarks: the similar scores on \texttt{SWE-Bench-Verified} and \texttt{SWE-Bench-Pro} indicate the signal is not specific to a single task distribution.

\Syntactic{} on \texttt{SWE-Bench-Verified} is a clear exception, with probes collapsing to near chance (AUC always below $0.60$).
This happens because trajectories in \texttt{SWE-Bench-Verified} contain mostly compilable Python programs (positive label rate above $0.92$) -- the considered models excel in this language.
In the case of \texttt{SWE-Bench-Pro}, which includes three additional programming languages in its tasks (e.g. C++, TypeScript), and a far richer mix of compilation failures (positive rate between $0.52$ and $0.57$) in our training datasets, we observe higher training data balance and higher signal (AUC up to $0.78$).

To sum up, as a coding agent edits a real codebase, its residual stream carries a linearly decodable representation of current program properties: whether it compiles, whether it passes its tests, and how its edits move the test suite relative to the initial repository.
Those results have strong external validity, based on 22714 trajectories collected over 1231 different coding tasks.

\begin{figure*}[tb]
  \centering
  \includegraphics[width=\linewidth]{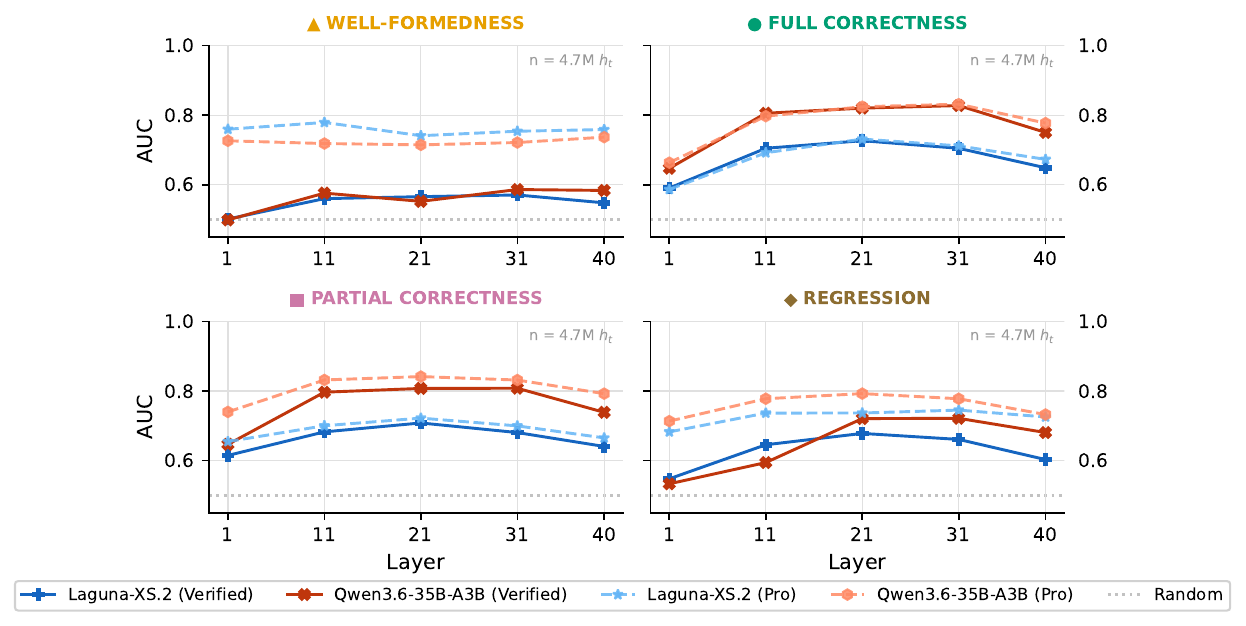}
  \caption{AUC-ROC by transformer layer for all four probes at $k=0$: \Syntactic{} (top left), \Semantic{} (top right), \RedFail{} (bottom left), \Regressions{} (bottom right). Each panel shows both models (\texttt{Laguna-XS.2} and \texttt{Qwen3.6-35B-A3B}) across both datasets (\texttt{SWE-Bench-Verified} and \texttt{SWE-Bench-Pro}), with separate lines per model--dataset combination. The inverted-U pattern, weakest at layer~1, peaking in the intermediate layers, slightly lower at the final layer, is consistent across all probes and both models except \Syntactic{}.}
  \label{fig:layer_auc}
\end{figure*}

\paragraph{Layer localization of program properties.}
\Cref{fig:layer_auc} shows a consistent inverted-U across layer depth.
Performance on program property prediction is lowest at the shallowest layer, rises to a peak at intermediate layers, and falls again slightly at the final layer.
This pattern holds across all probes, models, and datasets, all indicating that the program signal is most concentrated in the middle layers.

At layer~1, the probe is almost always the weakest of any depth, in some cases near or at the random baseline.
This is consistent with previous work in mechanistic interpretability outside the coding domain \citep{jin2025exploring}.
The mid-layer peaks match previous results on natural language tasks, showing that intermediate layers often encode richer representations \citep{skean2025layer}.
While better than the first layer, the final layer shows a slight drop in signal, suggesting that final layers trade abstract features for features closer to the next token prediction task.
Full per-layer results are in \cref{tab:transfer} in the appendix.

\paragraph{Model comparison.}
Per \Cref{fig:auc_barplot}, \texttt{Qwen3.6-35B-A3B} consistently encodes program properties more strongly than \texttt{Laguna-XS.2}.
The gap is roughly $0.10$ AUC on \Semantic{} and \RedFail{} across both benchmarks ($0.83$ vs.\ $0.73$ and $0.81$/$0.84$ vs.\ $0.71$/$0.72$), and roughly $0.04$--$0.05$ on \Regressions{}.
Crucially, the layer finding holds, with both models exhibiting an inverted-U peaking in the middle layers, and the final layer slightly below the peak.
The two models therefore differ in how linearly decodable the program representation is but not in where it is encoded.
This suggests that mechanistic interpretability tasks might be more appropriate on \texttt{Qwen3.6-35B-A3B} than with \texttt{Laguna-XS.2}.

\paragraph{Probes transfer across datasets.}

\begin{figure}[tb]
  \centering
  \includegraphics[width=\scfigwidth]{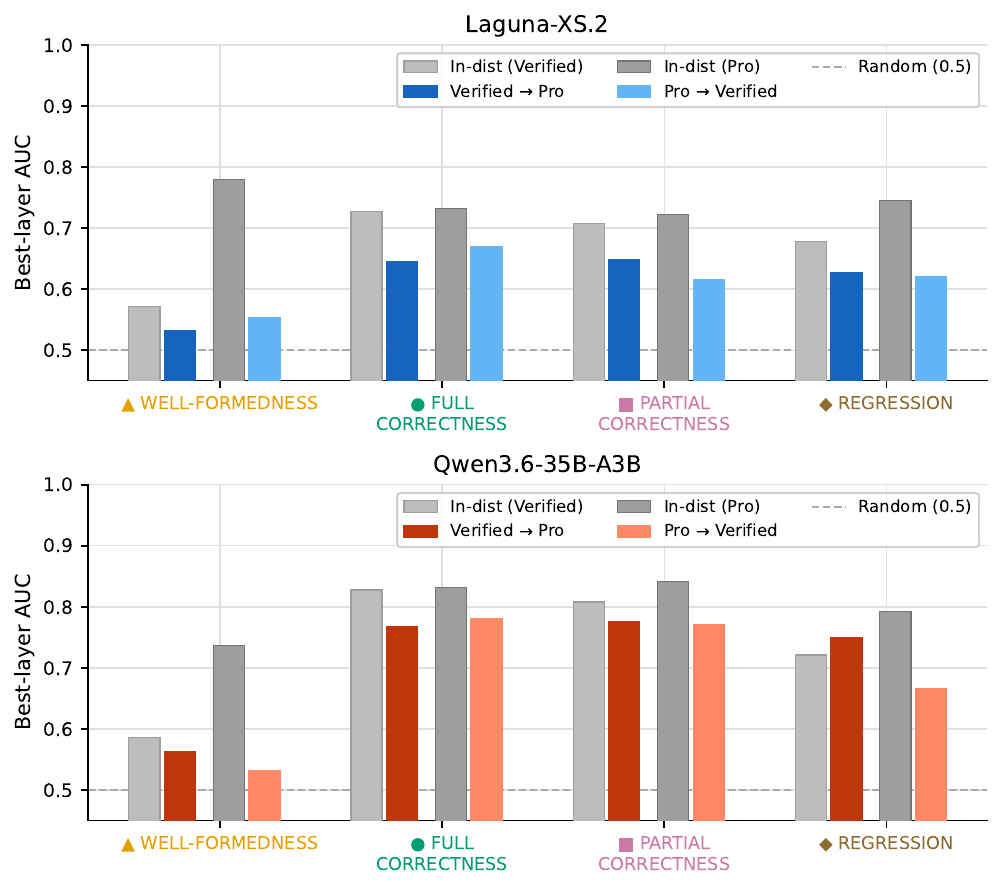}
  \caption{Cross-dataset transfer AUC at the best layer for each probe and model. Gray bars show in-distribution performance; colored bars show transfer performance when probe weights trained on one benchmark are applied to the other without retraining. \Semantic{} and \RedFail{} transfer with small drops for both models; \Syntactic{} collapses near chance in both directions due to the distributional mismatch in syntactic failure rates between \texttt{SWE-Bench-Verified} (near-constant compilability) and \texttt{SWE-Bench-Pro} (multi-language, richer failure modes).}
  \label{fig:transfer_barplot}
\end{figure}

\Cref{fig:transfer_barplot} tests whether the probes trained on one benchmark can, without retraining, retrieve program properties from trajectories on another dataset.
For both models, all semantic probes transfer substantially.
For example, \Semantic{} and \RedFail{} retain AUC of $0.63$--$0.78$ at the best layer under cross-dataset evaluation, compared to $0.71$--$0.84$ in-distribution, a drop of only $0.04$--$0.09$ units.
These results confirm that our probes extract real signal related to program properties from hidden states, adding further evidence to the hypothesis that coding agents maintain latent program representations.

\subsection{Coding agents have a long-term programming horizon}
\label{sec:results:lookahead}

\begin{figure*}[tb]
  \centering
  \begin{subfigure}[b]{0.48\textwidth}
    \includegraphics[width=\linewidth]{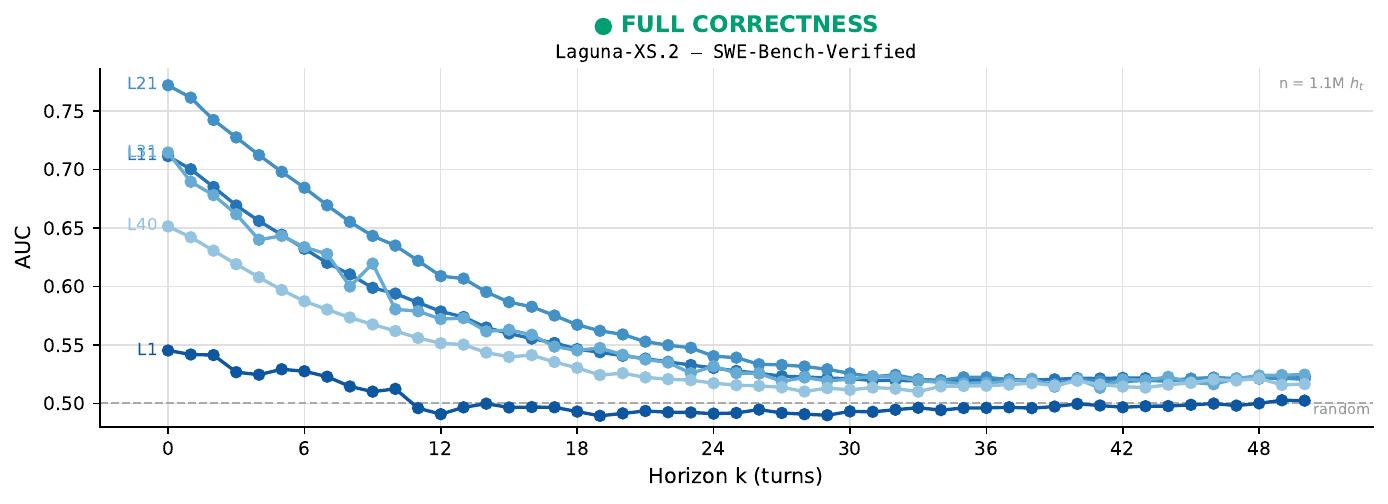}
  \end{subfigure}\hfill
  \begin{subfigure}[b]{0.48\textwidth}
    \includegraphics[width=\linewidth]{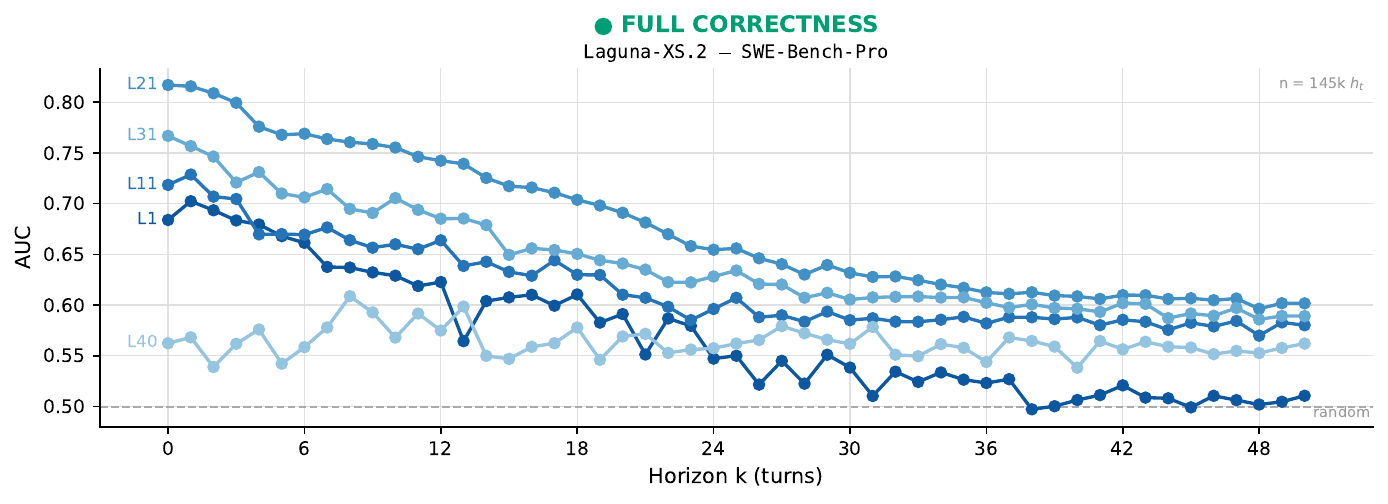}
  \end{subfigure}\\[0.5em]
  \begin{subfigure}[b]{0.48\textwidth}
    \includegraphics[width=\linewidth]{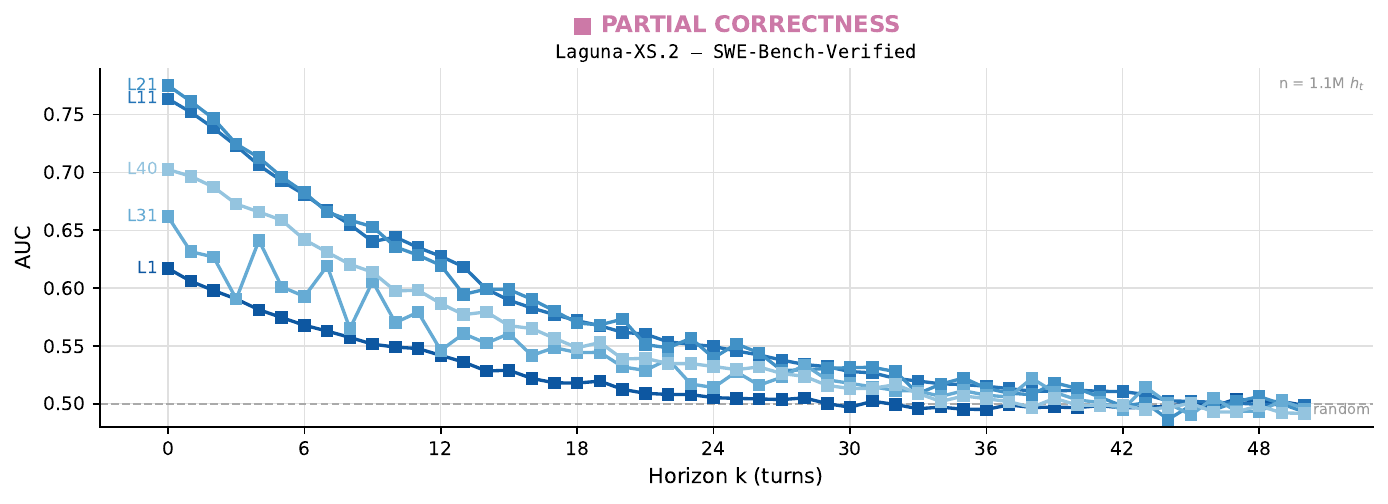}
  \end{subfigure}\hfill
  \begin{subfigure}[b]{0.48\textwidth}
    \includegraphics[width=\linewidth]{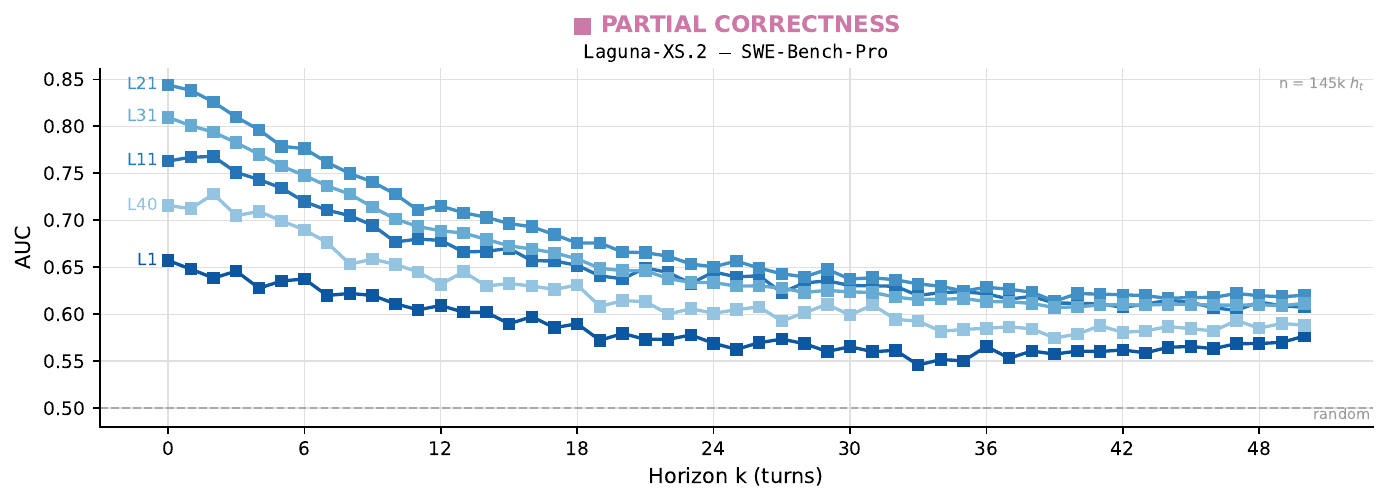}
  \end{subfigure}
  \caption{Lookahead AUC for the two strongest properties on \texttt{Laguna-XS.2}, across both benchmarks: \Semantic{} (top) and \RedFail{} (bottom); \texttt{SWE-Bench-Verified} (left) and \texttt{SWE-Bench-Pro} (right). A probe at token position $t$ predicts a program property $k$ steps later; the horizon $k$ runs from the edit itself ($k=0$, left of each plot) out to 50 steps ahead (right). Each line is a layer; the dashed line is the $0.50$ random baseline. AUC is highest at the edit, falls steeply over the first ${\sim}25$ steps, and plateaus above chance thereafter. The hockey-stick pattern is consistent across both benchmarks. Full per-probe results are in \cref{fig:lookahead_verified,fig:lookahead_pro} in the appendix.}
  \label{fig:lookahead}
\end{figure*}

We have shown that the hidden state encodes the current properties of the program.
Now, we demonstrate that it also encodes the properties of the future programs to come, that one can predict the outcome of future edits before those edits have materialized.
We train each probe to predict a program property $k$ steps in the future, $\hat{y}_{\sigma(t)+k}^\phi = \mathrm{probe}(\mathbf{h}_t^\ell)$, and sweep the horizon $k$ from $0$ (at the state to predict) out to $50$ steps ahead.
\Cref{fig:lookahead} plots the two strongest properties (\Semantic{} and \RedFail{}) on \texttt{Laguna-XS.2} across both benchmarks; each curve is a layer.

First, we see that the predictive signal decays smoothly with horizon (and drops most sharply after the initial edit which captures the present state of the program).
At the edit ($k=0$) the best-layer probe recovers \Semantic{} at AUC ${\sim}0.77$ on \texttt{SWE-Bench-Verified} and ${\sim}0.82$ on \texttt{SWE-Bench-Pro} (a slightly different figure than in \Cref{sec:results:current}, because this is a different test set).
For almost all layers, prediction remains above 0.5 for at least 25 steps.
As the horizon lengthens this falls, reaching ${\sim}0.55$ and ${\sim}0.65$ respectively around $k=25$, before flattening into a plateau that persists out to $k=50$.
The plateau actually sits above the $0.50$ control on both benchmarks (${\sim}0.52$ on Verified, ${\sim}0.60$ on Pro): even 50 steps ahead, an edit's eventual outcome is weakly predictable.
To our knowledge, this is the first evidence of long-term horizon by coding agents.

\section{Related Work}
\label{sec:related}

\paragraph{Probing internal representations.}
Probing, the method of training supervised classifiers on frozen hidden states to test for a target property, is a standard interpretability tool \citep{alain2017understanding, tenney2019bert, belinkov2022probing}.
For example, \citet{liemergent} probe a GPT-style model trained on Othello game transcripts and find an internal representation of the game board, which \citet{nanda2023emergent} show is recoverable with a linear probe.
Beyond game state, \citet{shai2024transformers} show that a transformer trained only on next-token prediction linearly encodes a belief-state geometry of the process that generated its training data.
\citet{park2024linear} give formal foundations to linear probing as a tool for interpreting internal representations.
\citet{elhage2022superposition} explain, through the theory of superposition, why many features remain linearly accessible even in high-dimensional residual streams.
The recommended best practice \citep{hewitt2019designing} is to use linear probes of limited capacity and compare every probe against a control to ensure the signal lies in the representation rather than the probe.

\paragraph{Predicting future states.}
Existing work finds that a single hidden state carries information about tokens not yet generated \citep{pal2023future, belrose2023eliciting, lindsey2025biology, pfaulet}.
At a higher level of abstraction, \citet{jenner2024chess} find that a chess-playing network represents future optimal moves, while \citet{taufeeque2024sokoban} show that a recurrent Sokoban agent encodes its planned actions up to roughly 50 steps ahead.
Closest to our correctness horizon, hidden states taken during chain-of-thought reasoning have been shown to encode the final answer and whether it will be correct.
\citet{zhangreasoning} and \citet{afzal2025knowing} decode answer correctness mid-reasoning, \citet{david2025temporal} track how this signal sharpens as reasoning proceeds, and \citet{cencerrado2025no} predict answer accuracy from a probe on the question alone, before any answer is generated.
These are all demonstrations of learned horizon capabilities.
We show the same for coding agents, where the decoded outcome is the result of an edit made many steps of tool use and test execution later.

\paragraph{Code representations in LLMs.}
A line of work probes code models for the structure they encode.
\citet{jin2024emergent} train a Transformer model on Karel-grid programs and show that linear probes on its hidden states recover both current and future program execution states.
\citet{wan2022they} find that attention patterns of BERT-style code models highly align with the syntactic structure of the input, and that an unlabeled binary constituency tree can be constructed from the models' hidden states in an unsupervised, training-free manner.
\citet{hernandez2022ast} further show that these models encode the full labeled abstract syntax tree of code input within a compact syntactic, orthogonally-extracted subspace of their token-level residual stream.
Moving toward semantics, \citet{karmakar2021pre, karmakar2023inspect} assemble a battery of probing tasks and report that code models capture semantic properties like type validity more readily as syntactic properties than structural ones such as cyclomatic complexity. \citet{troshin2022probing} probe a range of syntactic, namespace, data-flow and semantic properties, finding that code models encode syntactic structure and identifiers well but capture little about deep semantics such as computational equivalence. Ma et al. \citet{ma2024unveiling} separately measure how well models recover data-dependency, control-dependency and control-flow graphs across layers and additionally prove that decoder-only transformer, which is adopted by most LLMs today, possesses the same capability of extracting coding representation as BERT.

\paragraph{Correctness probing.} Closer to our target, a recent line probes hidden states for the correctness of generated programs.
\citet{spiess2025internal}, \citet{correctness2025internal}, and \citet{autoprobe2025} aim at decoding whether a single generated function is correct, the last by selecting the most informative layer per input; \citet{di2026code} finds correctness is linearly readable even before the code is generated.
\citet{he2026codecircuit} predict correctness from attribution graphs instead of a linear probe.
\citet{gros2025localized} use model calibration data to focus future patching edits towards a minimal repair.
\citet{tahimic2025mechanistic} use sparse autoencoders to find a correctness direction that is causally necessary for generating correct code, finding directions that predict errors but failing to do so for correctness.
All of this work studies a synthetic domain-specific language, isolated snippets, or a single step of generation. We instead probe models working on real-world programs, as coding agents rewrite them over many steps. To our knowledge, we are the first to study the outcome of future edits down the programming horizon.

\paragraph{Latent program spaces.}
A separate line of work explicitly constructs latent program spaces as part of the neural network modeling.
Early neural program induction represents and executes programs through latent operation embeddings, as in Neural Programmer-Interpreter \citep{reed2015neural} and Neural Programmer \citep{neelakantan2015neural}.
Subsequent methods build continuous program spaces and search them, for program synthesis \citep{hong2021latent, balog2020neural, macfarlane2026searching, macfarlanegradient} and for program repair \citep{silva2025gradient}.
We instead operate on off-the-shelf LLMs. We show, through linear probing, that language models trained to act as coding agents encode latent program representations. These representations have emerged as a byproduct of training, without any explicit latent space training objective.

\paragraph{Coding agents.}
Coding agents combine a language model with a harness that lets it autonomously edit and execute code to solve software engineering tasks, such as SWE-agent \citep{yang2024sweagent}, OpenHands \citep{wang2025openhands}, CodeAct \citep{wang2024executable}, and AutoCodeRover \citep{zhang2024autocoderover}.
Dedicated benchmarks evaluate them on real-world tasks, including SWE-bench \citep{jimenez2024swe}, SWE-bench Pro \citep{deng2025swe}, and Multi-SWE-bench \citep{zan2026multi}.
Analyses of agent trajectories  \citep{bouzenia2025understanding} show that successful and failed runs are structurally distinguishable in the token space, with failed trajectories being consistently longer and more variable in step and token count.
Like us, \citet{sui2026tact} probe the residual stream of coding agents for early detection of failures. They do not study the properties of programs in the latent program space.
To the best of our knowledge, we are the first to demonstrate that coding agents maintain a latent representation of the program throughout their past changes and future editing horizon.

\section{Limitations}
\label{sec:limitations}

\paragraph{Decodability is not causality.}
Our probes establish that program properties are linearly decodable from hidden states; they do not establish that the agent uses this information when surfacing actions.
Our claims are about what is represented, not about causal mechanisms.
Demonstrating causality would require steering along a probe direction and measuring the effect on the resulting edit and we leave it to future work.

\paragraph{Effect of label imbalance.}
\Semantic{} and \RedFail{} probes achieve the highest AUC across both models, while \Syntactic{} consistently achieves the lowest.
We attribute this to label imbalance.
\Syntactic{} is overwhelmingly positive in the collected trajectories because today's models excel at producing syntactically valid programs. 
The same imbalance explains why the Brier score for \Syntactic{} is low despite its low AUC, since a near-constant predictor scores well on Brier by matching the label base rate. Overall, it is most interesting to predict the hardest program properties, i.e., the ones that often flip and take the longest to achieve. 

\paragraph{Scope.}
We study two open-weight models under a single agent scaffold (\texttt{mini-swe-agent}) on two benchmarks.
We believe that the same representation evidence and horizons appear in frontier models, under different scaffolds, or on tasks beyond the ones present in these datasets, but this remains to be demonstrated empirically.

\section{Conclusion}
\label{sec:conclusion}

Coding agents encode properties of the program they are working on in their residual streams.
We demonstrated that linear probes trained on hidden states decode whether the code compiles, passes its test suite, reduces failing tests, and introduces regressions, reaching AUC up to $0.83$ for full correctness.
This latent representation captures both the present program properties, but also can ground predictions on what the agent has not yet written:
probes predicting the outcome of future edits succeed above chance up to 25 steps before the edit is made.
These results have external validity, they hold across two models and two benchmarks and transfer across datasets without retraining.
These findings open novel research directions towards monitoring and steering coding agents from within the latent space.

\subsubsection*{Acknowledgments}
This work was partially supported by the Wallenberg AI, Autonomous Systems and Software Program (WASP) funded by the Knut and Alice Wallenberg Foundation.
Computational resources were provided on the Berzelius system funded by the Knut and Alice Wallenberg foundation and operated by NAISS.
We acknowledge compute support from Modal.

\ifcolm
  \bibliographystyle{colm2026_conference}
\fi
\bibliography{main}

\appendix
\onecolumn

\section{Appendix}

\subsection{Dataset Statistics}
\label{app:dataset_stats}

\begin{figure}[h]
  \centering
  \begin{subfigure}[b]{0.32\textwidth}
    \includegraphics[width=\linewidth]{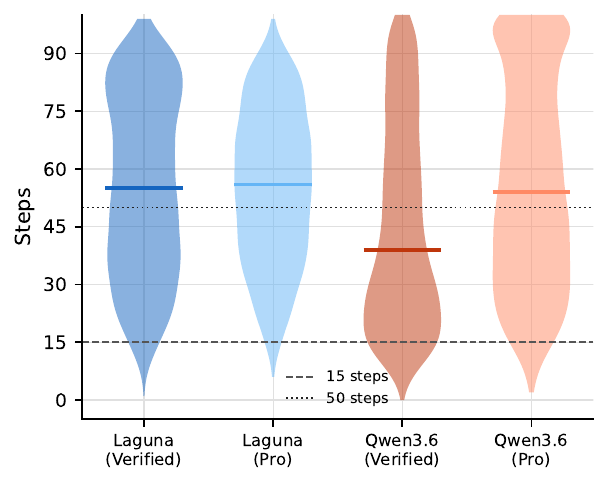}
  \end{subfigure}\hfill
  \begin{subfigure}[b]{0.32\textwidth}
    \includegraphics[width=\linewidth]{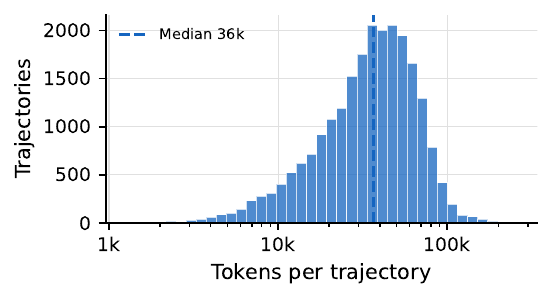}
  \end{subfigure}\hfill
  \begin{subfigure}[b]{0.32\textwidth}
    \includegraphics[width=\linewidth]{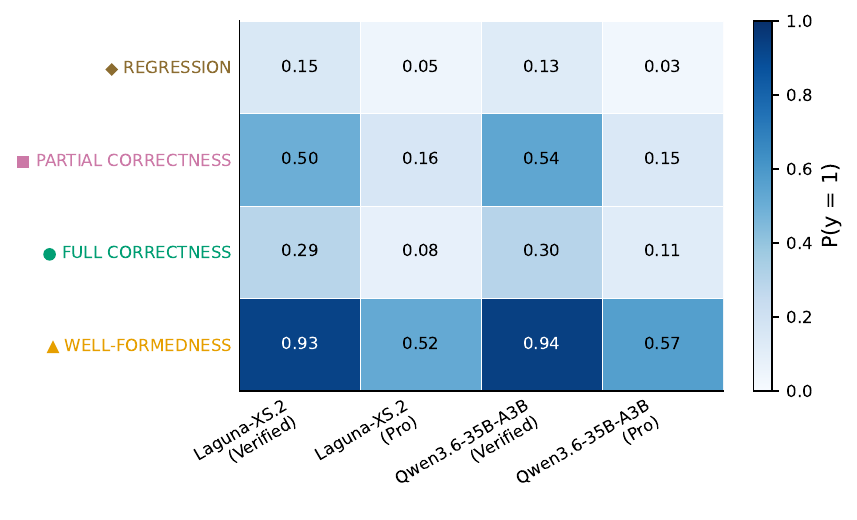}
  \end{subfigure}
  \caption{Left: distribution of trajectory lengths (in agent steps) per (model, dataset) combination. Dashed vertical lines mark the $k_{\max}=15$ and $k_{\max}=50$ lookahead thresholds; the former is nearly lossless (90--99\% of trajectories), while the latter retains 40--60\%. Middle: distribution of tokens per trajectory (median 36k). Right: positive label rate $P(y=1)$ per probe and (model, dataset) combination. \Syntactic{} is nearly always positive, which suppresses its AUC despite a real signal being present (see Limitations); \Semantic{} and \RedFail{} are more balanced, consistent with their higher AUC.}
  \label{fig:dataset_stats}
\end{figure}

\begin{table}[h]
\centering
\footnotesize
\caption{Agent hyperparameters used during trajectory generation, following the recommended settings provided by each model vendor. Temperature and top-$p$ control sampling; context is the maximum sequence length; max steps is the step limit per trajectory; and step timeout is the wall-clock limit per tool call.}
\label{tab:agent_hparams}
\begin{tabular}{lrrrrr}
\toprule
Model & Temp. & Top-$p$ & Context & Max steps & Step timeout \\
\midrule
\texttt{Laguna-XS.2} & 0.7 & 0.95 & 256k & 100 & 180\,s \\
\texttt{Qwen3.6-35B-A3B} & 0.6 & 0.95 & 256k & 100 & 180\,s \\
\bottomrule
\end{tabular}
\end{table}

\subsection{Probe Hyperparameters}
\label{app:probetrain}

\begin{table*}[h]
  \centering
  \begin{tabular}{lll}
  \toprule
  \textbf{Hyperparameter} & \textbf{Distribution} & \textbf{Values} \\
  \midrule
  Learning rate       & Log-uniform  & $[10^{-4},\ 10^{-1}]$ \\
  Weight decay        & Log-uniform  & $[10^{-5},\ 10^{-1}]$ \\
  Batch size          & Categorical  & $\{256,\ 512,\ 1024\}$ \\
  Patience & Categorical & $\{10,\ 25,\ 50\}$ \\
  \bottomrule
  \end{tabular}
  \caption{Hyperparameter search space used in the random search sweeps with 20 trials each. The best configurations, maximising validation AUC per probe per layer per model per dataset are used for training the final probes.}
  \label{tab:hp-sweep}
\end{table*}

\begin{table}[t]
\centering
\caption{Chosen hyperparameters for each model and benchmark, selected by 20-trial random search maximising mean validation AUC. Sweeps were run independently per probe and per layer; table shows the middle probed layer as representative.}
\label{tab:probe-hparams}
\resizebox{\columnwidth}{!}{%
\begin{tabular}{lllrrrr}
\toprule
Dataset & Model & Probe & Learning rate & Weight decay & Batch size & Patience \\
\midrule
\multirow{8}{*}{\textsc{Verified}} & \multirow{4}{*}{Laguna-XS.2} & \Syntactic{} & $8.58 \times 10^{-3}$ & $1.18 \times 10^{-3}$ & 512 & 50 \\
 &  & \Semantic{} & $5.19 \times 10^{-4}$ & $3.81 \times 10^{-3}$ & 512 & 10 \\
 &  & \Regressions{} & $2.95 \times 10^{-2}$ & $5.19 \times 10^{-5}$ & 256 & 50 \\
 &  & \RedFail{} & $2.66 \times 10^{-2}$ & $2.13 \times 10^{-3}$ & 1024 & 50 \\
 & \multirow{4}{*}{Qwen3.6-35B-A3B} & \Syntactic{} & $3.33 \times 10^{-3}$ & $4.20 \times 10^{-2}$ & 512 & 25 \\
 &  & \Semantic{} & $1.17 \times 10^{-2}$ & $4.10 \times 10^{-4}$ & 1024 & 25 \\
 &  & \Regressions{} & $3.11 \times 10^{-4}$ & $1.91 \times 10^{-4}$ & 512 & 50 \\
 &  & \RedFail{} & $3.39 \times 10^{-3}$ & $1.75 \times 10^{-4}$ & 1024 & 10 \\
\midrule
\multirow{8}{*}{\textsc{Pro}} & \multirow{4}{*}{Laguna-XS.2} & \Syntactic{} & $1.40 \times 10^{-3}$ & $3.93 \times 10^{-4}$ & 512 & 25 \\
 &  & \Semantic{} & $2.09 \times 10^{-4}$ & $6.05 \times 10^{-4}$ & 256 & 25 \\
 &  & \RedFail{} & $1.52 \times 10^{-3}$ & $1.64 \times 10^{-3}$ & 1024 & 50 \\
 &  & \Regressions{} & $1.13 \times 10^{-4}$ & $1.49 \times 10^{-2}$ & 1024 & 10 \\
 & \multirow{4}{*}{Qwen3.6-35B-A3B} & \Syntactic{} & $2.30 \times 10^{-4}$ & $1.46 \times 10^{-5}$ & 512 & 10 \\
 &  & \Semantic{} & $4.15 \times 10^{-4}$ & $2.27 \times 10^{-4}$ & 1024 & 10 \\
 &  & \Regressions{} & $2.74 \times 10^{-3}$ & $2.00 \times 10^{-4}$ & 512 & 50 \\
 &  & \RedFail{} & $1.26 \times 10^{-3}$ & $8.97 \times 10^{-5}$ & 1024 & 50 \\
\bottomrule
\end{tabular}%
}
\end{table}

\subsection{Full Probe Results}
\label{app:auc}

\begin{table*}[t]
\centering
\caption{AUC per probe, model, and benchmark across transformer layers. \textbf{Bold} marks the best layer per row. \emph{Shuffled} uses label-permuted data as a sanity baseline.}
\label{tab:auc_roc}
\begin{tabular}{lcccccc}
\toprule
 & \multicolumn{5}{c}{AUC $\uparrow$} & Shuffled \\
\cmidrule(lr){2-6}
Probe & 1 & 11 & 21 & 31 & 40 & (best layer) \\
\midrule
\multicolumn{7}{l}{\textsc{SWE-bench Verified}} \\
\addlinespace[2pt]
\multicolumn{7}{l}{\quad\textit{Laguna-XS.2}} \\
\Syntactic{} & 0.502 & 0.560 & 0.566 & \textbf{0.571} & 0.548 & 0.500 \\
\Semantic{} & 0.591 & 0.705 & \textbf{0.727} & 0.705 & 0.649 & 0.501 \\
\RedFail{} & 0.615 & 0.682 & \textbf{0.708} & 0.680 & 0.641 & 0.500 \\
\Regressions{} & 0.548 & 0.645 & \textbf{0.678} & 0.661 & 0.603 & 0.500 \\
\addlinespace[3pt]
\multicolumn{7}{l}{\quad\textit{Qwen3.6-35B-A3B}} \\
\Syntactic{} & 0.499 & 0.576 & 0.553 & \textbf{0.587} & 0.584 & 0.501 \\
\Semantic{} & 0.648 & 0.806 & 0.820 & \textbf{0.828} & 0.751 & 0.501 \\
\RedFail{} & 0.646 & 0.796 & 0.807 & \textbf{0.808} & 0.739 & 0.501 \\
\Regressions{} & 0.533 & 0.594 & 0.721 & \textbf{0.721} & 0.681 & 0.501 \\
\midrule
\multicolumn{7}{l}{\textsc{SWE-bench Pro}} \\
\addlinespace[2pt]
\multicolumn{7}{l}{\quad\textit{Laguna-XS.2}} \\
\Syntactic{} & 0.760 & \textbf{0.779} & 0.741 & 0.754 & 0.759 & 0.501 \\
\Semantic{} & 0.587 & 0.692 & \textbf{0.732} & 0.712 & 0.673 & 0.500 \\
\RedFail{} & 0.655 & 0.700 & \textbf{0.722} & 0.700 & 0.665 & 0.500 \\
\Regressions{} & 0.682 & 0.736 & 0.737 & \textbf{0.745} & 0.725 & 0.501 \\
\addlinespace[3pt]
\multicolumn{7}{l}{\quad\textit{Qwen3.6-35B-A3B}} \\
\Syntactic{} & 0.727 & 0.719 & 0.715 & 0.722 & \textbf{0.737} & 0.502 \\
\Semantic{} & 0.664 & 0.797 & 0.825 & \textbf{0.832} & 0.777 & 0.502 \\
\RedFail{} & 0.740 & 0.832 & \textbf{0.841} & 0.831 & 0.792 & 0.502 \\
\Regressions{} & 0.713 & 0.778 & \textbf{0.792} & 0.778 & 0.732 & 0.507 \\
\bottomrule
\end{tabular}
\end{table*}

\begin{table*}[t]
\centering
\caption{Cross-dataset transfer AUC. Gray rows show in-distribution reference performance. Transfer rows show AUC when probe weights trained on one dataset are evaluated on the other; the subscript shows the delta relative to the in-distribution baseline on the \emph{same} evaluation set.}
\label{tab:transfer}
\begin{tabular}{lccccc}
\toprule
 & Layer 1 & Layer 11 & Layer 21 & Layer 31 & Layer 40 \\
\midrule
\multicolumn{6}{l}{\textit{\Syntactic{}}} \\
\quad{}\textit{Laguna-XS.2} \\
\qquad{}\textcolor{gray}{\small In-dist (Verified)} & \textcolor{gray}{\small 0.502} & \textcolor{gray}{\small 0.560} & \textcolor{gray}{\small 0.566} & \textcolor{gray}{\small 0.571} & \textcolor{gray}{\small 0.548} \\
\qquad{}\textcolor{gray}{\small In-dist (Pro)} & \textcolor{gray}{\small 0.760} & \textcolor{gray}{\small 0.779} & \textcolor{gray}{\small 0.741} & \textcolor{gray}{\small 0.754} & \textcolor{gray}{\small 0.759} \\
\qquad{}Verified $\rightarrow$ Pro & 0.508\,{\scriptsize \textcolor{red!70!black}{$-0.251$}} & 0.527\,{\scriptsize \textcolor{red!70!black}{$-0.252$}} & 0.532\,{\scriptsize \textcolor{red!70!black}{$-0.210$}} & 0.523\,{\scriptsize \textcolor{red!70!black}{$-0.231$}} & 0.508\,{\scriptsize \textcolor{red!70!black}{$-0.251$}}  \\
\qquad{}Pro $\rightarrow$ Verified & 0.484\,{\scriptsize \textcolor{red!70!black}{$-0.018$}} & 0.521\,{\scriptsize \textcolor{red!70!black}{$-0.040$}} & 0.553\,{\scriptsize \textcolor{red!70!black}{$-0.013$}} & 0.528\,{\scriptsize \textcolor{red!70!black}{$-0.043$}} & 0.498\,{\scriptsize \textcolor{red!70!black}{$-0.050$}}  \\
\quad{}\textit{Qwen3.6-35B-A3B} \\
\qquad{}\textcolor{gray}{\small In-dist (Verified)} & \textcolor{gray}{\small 0.499} & \textcolor{gray}{\small 0.576} & \textcolor{gray}{\small 0.553} & \textcolor{gray}{\small 0.587} & \textcolor{gray}{\small 0.584} \\
\qquad{}\textcolor{gray}{\small In-dist (Pro)} & \textcolor{gray}{\small 0.727} & \textcolor{gray}{\small 0.719} & \textcolor{gray}{\small 0.715} & \textcolor{gray}{\small 0.722} & \textcolor{gray}{\small 0.737} \\
\qquad{}Verified $\rightarrow$ Pro & 0.564\,{\scriptsize \textcolor{red!70!black}{$-0.163$}} & 0.516\,{\scriptsize \textcolor{red!70!black}{$-0.202$}} & 0.516\,{\scriptsize \textcolor{red!70!black}{$-0.199$}} & 0.516\,{\scriptsize \textcolor{red!70!black}{$-0.206$}} & 0.480\,{\scriptsize \textcolor{red!70!black}{$-0.257$}}  \\
\qquad{}Pro $\rightarrow$ Verified & 0.509\,{\scriptsize \textcolor{green!50!black}{$+0.010$}} & 0.527\,{\scriptsize \textcolor{red!70!black}{$-0.049$}} & 0.531\,{\scriptsize \textcolor{red!70!black}{$-0.021$}} & 0.503\,{\scriptsize \textcolor{red!70!black}{$-0.083$}} & 0.447\,{\scriptsize \textcolor{red!70!black}{$-0.137$}}  \\
\midrule
\multicolumn{6}{l}{\textit{\Semantic{}}} \\
\quad{}\textit{Laguna-XS.2} \\
\qquad{}\textcolor{gray}{\small In-dist (Verified)} & \textcolor{gray}{\small 0.591} & \textcolor{gray}{\small 0.705} & \textcolor{gray}{\small 0.727} & \textcolor{gray}{\small 0.705} & \textcolor{gray}{\small 0.649} \\
\qquad{}\textcolor{gray}{\small In-dist (Pro)} & \textcolor{gray}{\small 0.587} & \textcolor{gray}{\small 0.692} & \textcolor{gray}{\small 0.732} & \textcolor{gray}{\small 0.712} & \textcolor{gray}{\small 0.673} \\
\qquad{}Verified $\rightarrow$ Pro & 0.502\,{\scriptsize \textcolor{red!70!black}{$-0.085$}} & 0.630\,{\scriptsize \textcolor{red!70!black}{$-0.062$}} & 0.645\,{\scriptsize \textcolor{red!70!black}{$-0.086$}} & 0.643\,{\scriptsize \textcolor{red!70!black}{$-0.069$}} & 0.542\,{\scriptsize \textcolor{red!70!black}{$-0.132$}}  \\
\qquad{}Pro $\rightarrow$ Verified & 0.531\,{\scriptsize \textcolor{red!70!black}{$-0.060$}} & 0.634\,{\scriptsize \textcolor{red!70!black}{$-0.071$}} & 0.669\,{\scriptsize \textcolor{red!70!black}{$-0.057$}} & 0.636\,{\scriptsize \textcolor{red!70!black}{$-0.069$}} & 0.594\,{\scriptsize \textcolor{red!70!black}{$-0.055$}}  \\
\quad{}\textit{Qwen3.6-35B-A3B} \\
\qquad{}\textcolor{gray}{\small In-dist (Verified)} & \textcolor{gray}{\small 0.648} & \textcolor{gray}{\small 0.806} & \textcolor{gray}{\small 0.820} & \textcolor{gray}{\small 0.828} & \textcolor{gray}{\small 0.751} \\
\qquad{}\textcolor{gray}{\small In-dist (Pro)} & \textcolor{gray}{\small 0.664} & \textcolor{gray}{\small 0.797} & \textcolor{gray}{\small 0.825} & \textcolor{gray}{\small 0.832} & \textcolor{gray}{\small 0.777} \\
\qquad{}Verified $\rightarrow$ Pro & 0.615\,{\scriptsize \textcolor{red!70!black}{$-0.048$}} & 0.730\,{\scriptsize \textcolor{red!70!black}{$-0.068$}} & 0.748\,{\scriptsize \textcolor{red!70!black}{$-0.076$}} & 0.767\,{\scriptsize \textcolor{red!70!black}{$-0.064$}} & 0.702\,{\scriptsize \textcolor{red!70!black}{$-0.076$}}  \\
\qquad{}Pro $\rightarrow$ Verified & 0.587\,{\scriptsize \textcolor{red!70!black}{$-0.061$}} & 0.764\,{\scriptsize \textcolor{red!70!black}{$-0.042$}} & 0.781\,{\scriptsize \textcolor{red!70!black}{$-0.039$}} & 0.778\,{\scriptsize \textcolor{red!70!black}{$-0.049$}} & 0.699\,{\scriptsize \textcolor{red!70!black}{$-0.052$}}  \\
\midrule
\multicolumn{6}{l}{\textit{\RedFail{}}} \\
\quad{}\textit{Laguna-XS.2} \\
\qquad{}\textcolor{gray}{\small In-dist (Verified)} & \textcolor{gray}{\small 0.615} & \textcolor{gray}{\small 0.682} & \textcolor{gray}{\small 0.708} & \textcolor{gray}{\small 0.680} & \textcolor{gray}{\small 0.641} \\
\qquad{}\textcolor{gray}{\small In-dist (Pro)} & \textcolor{gray}{\small 0.655} & \textcolor{gray}{\small 0.700} & \textcolor{gray}{\small 0.722} & \textcolor{gray}{\small 0.700} & \textcolor{gray}{\small 0.665} \\
\qquad{}Verified $\rightarrow$ Pro & 0.519\,{\scriptsize \textcolor{red!70!black}{$-0.136$}} & 0.622\,{\scriptsize \textcolor{red!70!black}{$-0.078$}} & 0.648\,{\scriptsize \textcolor{red!70!black}{$-0.074$}} & 0.625\,{\scriptsize \textcolor{red!70!black}{$-0.075$}} & 0.556\,{\scriptsize \textcolor{red!70!black}{$-0.109$}}  \\
\qquad{}Pro $\rightarrow$ Verified & 0.517\,{\scriptsize \textcolor{red!70!black}{$-0.097$}} & 0.574\,{\scriptsize \textcolor{red!70!black}{$-0.109$}} & 0.615\,{\scriptsize \textcolor{red!70!black}{$-0.093$}} & 0.577\,{\scriptsize \textcolor{red!70!black}{$-0.104$}} & 0.551\,{\scriptsize \textcolor{red!70!black}{$-0.090$}}  \\
\quad{}\textit{Qwen3.6-35B-A3B} \\
\qquad{}\textcolor{gray}{\small In-dist (Verified)} & \textcolor{gray}{\small 0.646} & \textcolor{gray}{\small 0.796} & \textcolor{gray}{\small 0.807} & \textcolor{gray}{\small 0.808} & \textcolor{gray}{\small 0.739} \\
\qquad{}\textcolor{gray}{\small In-dist (Pro)} & \textcolor{gray}{\small 0.740} & \textcolor{gray}{\small 0.832} & \textcolor{gray}{\small 0.841} & \textcolor{gray}{\small 0.831} & \textcolor{gray}{\small 0.792} \\
\qquad{}Verified $\rightarrow$ Pro & 0.623\,{\scriptsize \textcolor{red!70!black}{$-0.117$}} & 0.774\,{\scriptsize \textcolor{red!70!black}{$-0.058$}} & 0.773\,{\scriptsize \textcolor{red!70!black}{$-0.068$}} & 0.776\,{\scriptsize \textcolor{red!70!black}{$-0.055$}} & 0.720\,{\scriptsize \textcolor{red!70!black}{$-0.072$}}  \\
\qquad{}Pro $\rightarrow$ Verified & 0.605\,{\scriptsize \textcolor{red!70!black}{$-0.041$}} & 0.754\,{\scriptsize \textcolor{red!70!black}{$-0.042$}} & 0.766\,{\scriptsize \textcolor{red!70!black}{$-0.041$}} & 0.771\,{\scriptsize \textcolor{red!70!black}{$-0.037$}} & 0.699\,{\scriptsize \textcolor{red!70!black}{$-0.040$}}  \\
\midrule
\multicolumn{6}{l}{\textit{\Regressions{}}} \\
\quad{}\textit{Laguna-XS.2} \\
\qquad{}\textcolor{gray}{\small In-dist (Verified)} & \textcolor{gray}{\small 0.548} & \textcolor{gray}{\small 0.645} & \textcolor{gray}{\small 0.678} & \textcolor{gray}{\small 0.661} & \textcolor{gray}{\small 0.603} \\
\qquad{}\textcolor{gray}{\small In-dist (Pro)} & \textcolor{gray}{\small 0.682} & \textcolor{gray}{\small 0.736} & \textcolor{gray}{\small 0.737} & \textcolor{gray}{\small 0.745} & \textcolor{gray}{\small 0.725} \\
\qquad{}Verified $\rightarrow$ Pro & 0.459\,{\scriptsize \textcolor{red!70!black}{$-0.223$}} & 0.500\,{\scriptsize \textcolor{red!70!black}{$-0.237$}} & 0.626\,{\scriptsize \textcolor{red!70!black}{$-0.110$}} & 0.618\,{\scriptsize \textcolor{red!70!black}{$-0.127$}} & 0.540\,{\scriptsize \textcolor{red!70!black}{$-0.185$}}  \\
\qquad{}Pro $\rightarrow$ Verified & 0.560\,{\scriptsize \textcolor{green!50!black}{$+0.012$}} & 0.612\,{\scriptsize \textcolor{red!70!black}{$-0.033$}} & 0.620\,{\scriptsize \textcolor{red!70!black}{$-0.058$}} & 0.601\,{\scriptsize \textcolor{red!70!black}{$-0.059$}} & 0.560\,{\scriptsize \textcolor{red!70!black}{$-0.043$}}  \\
\quad{}\textit{Qwen3.6-35B-A3B} \\
\qquad{}\textcolor{gray}{\small In-dist (Verified)} & \textcolor{gray}{\small 0.533} & \textcolor{gray}{\small 0.594} & \textcolor{gray}{\small 0.721} & \textcolor{gray}{\small 0.721} & \textcolor{gray}{\small 0.681} \\
\qquad{}\textcolor{gray}{\small In-dist (Pro)} & \textcolor{gray}{\small 0.713} & \textcolor{gray}{\small 0.778} & \textcolor{gray}{\small 0.792} & \textcolor{gray}{\small 0.778} & \textcolor{gray}{\small 0.732} \\
\qquad{}Verified $\rightarrow$ Pro & 0.575\,{\scriptsize \textcolor{red!70!black}{$-0.139$}} & 0.593\,{\scriptsize \textcolor{red!70!black}{$-0.185$}} & 0.750\,{\scriptsize \textcolor{red!70!black}{$-0.042$}} & 0.724\,{\scriptsize \textcolor{red!70!black}{$-0.053$}} & 0.641\,{\scriptsize \textcolor{red!70!black}{$-0.091$}}  \\
\qquad{}Pro $\rightarrow$ Verified & 0.557\,{\scriptsize \textcolor{green!50!black}{$+0.024$}} & 0.664\,{\scriptsize \textcolor{green!50!black}{$+0.070$}} & 0.667\,{\scriptsize \textcolor{red!70!black}{$-0.054$}} & 0.661\,{\scriptsize \textcolor{red!70!black}{$-0.060$}} & 0.613\,{\scriptsize \textcolor{red!70!black}{$-0.067$}}  \\
\bottomrule
\end{tabular}
\end{table*}

\subsection{Lookahead Analysis}
\label{app:lookahead}

The main lookahead experiment uses a maximum $k$ of 50, which excludes tokens within 50 steps of the end of a trajectory, including all trajectories shorter than 50 steps.
\Cref{fig:lookahead_max15} replicates the analysis at $k \leq 15$, which imposes a smaller constraint.
The monotonic rise in AUC as $k \to 0$ and the layer-ordering are consistent with the main results, confirming that the hockey-stick pattern is not an artifact of the reduced sample.

\begin{figure*}[h]
  \centering
  \begin{subfigure}[b]{0.48\textwidth}
    \includegraphics[width=\linewidth]{figures/figures/laguna_xs2_full_pooled/currently_correct_lookahead_max50.pdf}
  \end{subfigure}\hfill
  \begin{subfigure}[b]{0.48\textwidth}
    \includegraphics[width=\linewidth]{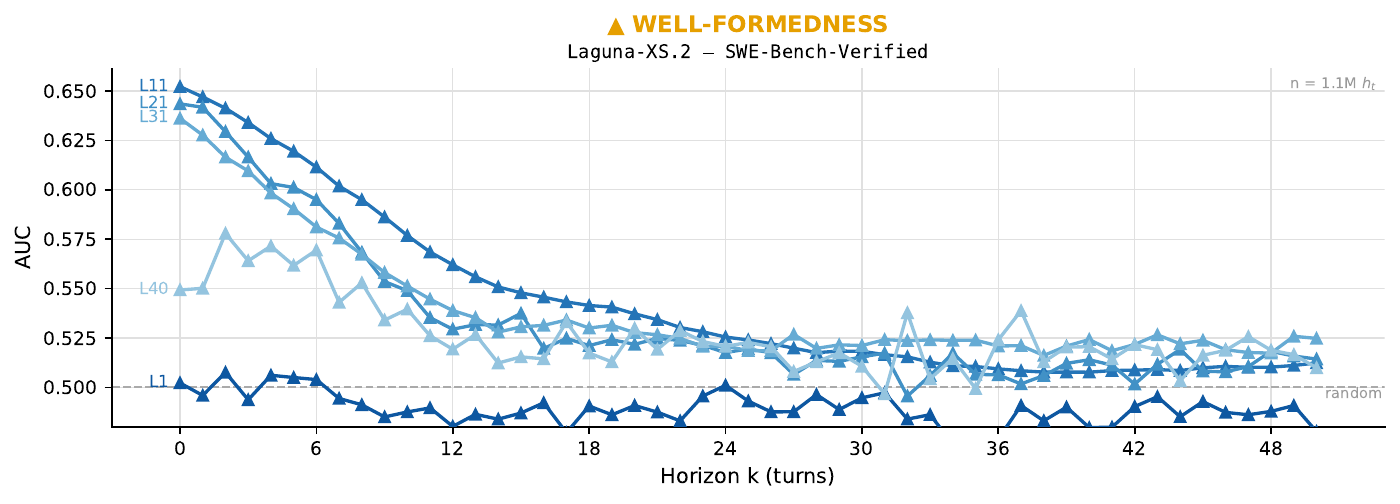}
  \end{subfigure}\\[0.5em]
  \begin{subfigure}[b]{0.48\textwidth}
    \includegraphics[width=\linewidth]{figures/figures/laguna_xs2_full_pooled/currently_reduces_failing_lookahead_max50.pdf}
  \end{subfigure}\hfill
  \begin{subfigure}[b]{0.48\textwidth}
    \includegraphics[width=\linewidth]{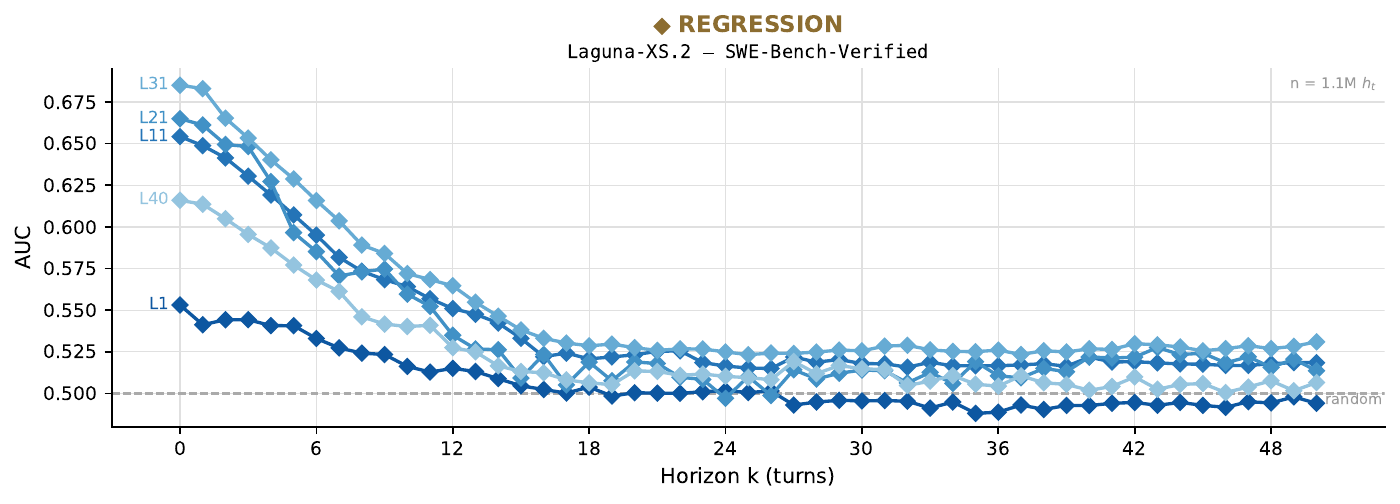}
  \end{subfigure}
  \caption{Lookahead AUC for all four probes on \texttt{Laguna-XS.2} (\texttt{SWE-Bench-Verified}), $k \leq 50$. Panels show \Semantic{} (top left), \Syntactic{} (top right), \RedFail{} (bottom left), and \Regressions{} (bottom right). A probe at token position $t$ predicts the program property $k$ steps later; each line is a transformer layer; the dashed line marks the $0.50$ random baseline. AUC rises as $k \to 0$ across all probes.}
  \label{fig:lookahead_verified}
\end{figure*}

\begin{figure*}[h]
  \centering
  \begin{subfigure}[b]{0.48\textwidth}
    \includegraphics[width=\linewidth]{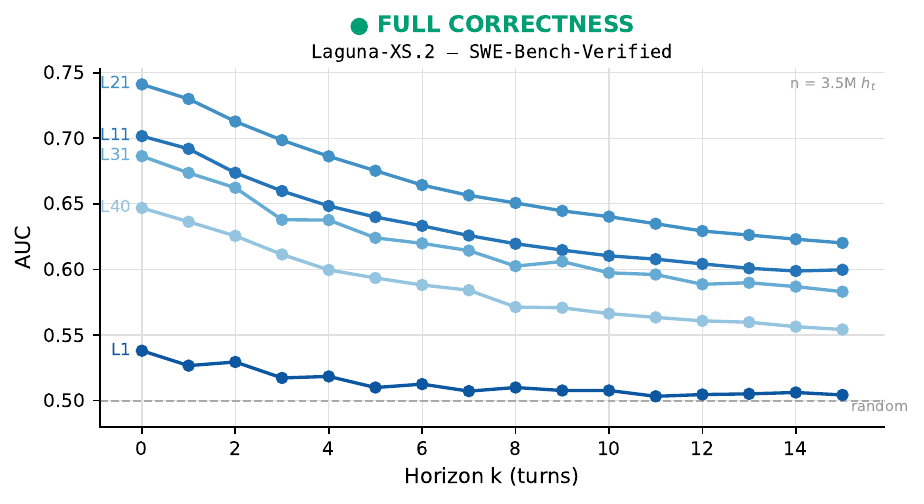}
  \end{subfigure}\hfill
  \begin{subfigure}[b]{0.48\textwidth}
    \includegraphics[width=\linewidth]{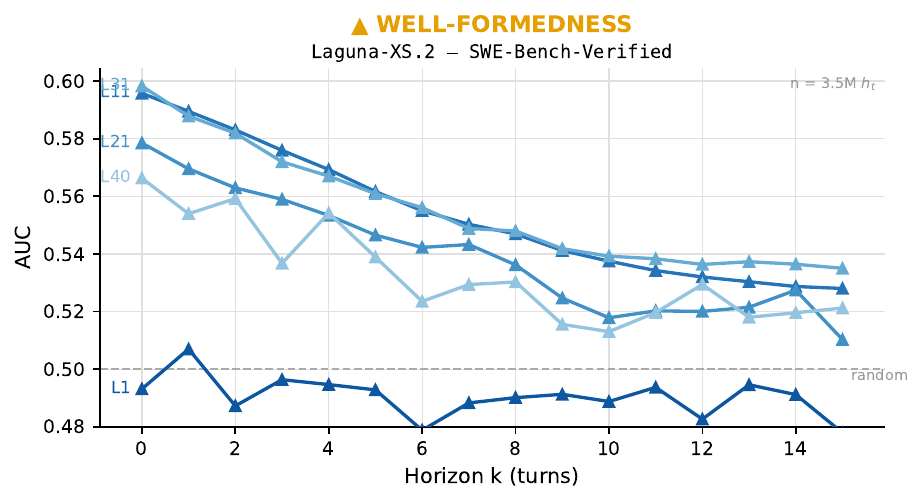}
  \end{subfigure}\\[0.5em]
  \begin{subfigure}[b]{0.48\textwidth}
    \includegraphics[width=\linewidth]{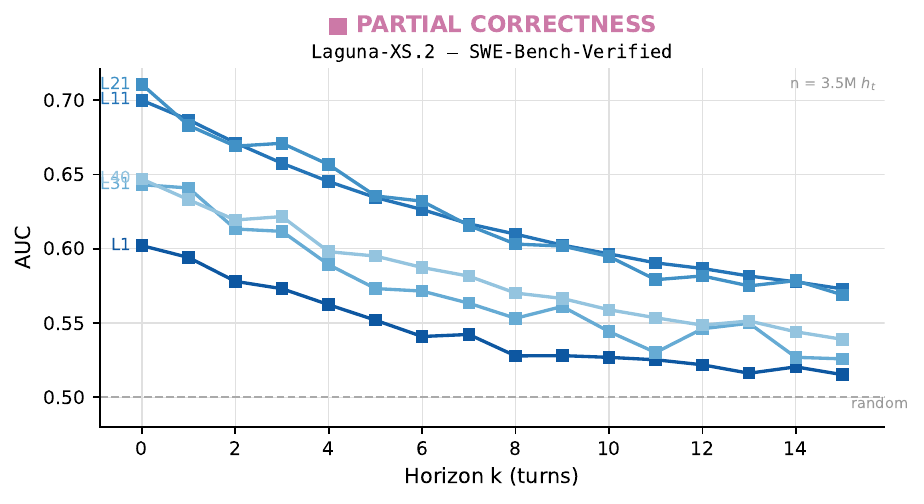}
  \end{subfigure}\hfill
  \begin{subfigure}[b]{0.48\textwidth}
    \includegraphics[width=\linewidth]{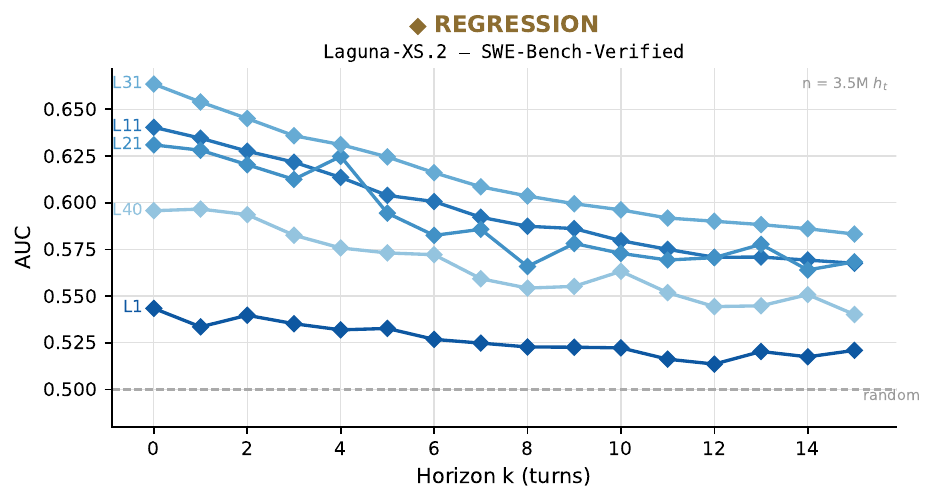}
  \end{subfigure}
  \caption{Lookahead AUC for all four probes on \texttt{Laguna-XS.2} (\texttt{SWE-Bench-Verified}), $k \leq 15$, using the full trajectory sample without the length filter required for $k \leq 50$. Panel layout matches \Cref{fig:lookahead_verified}. The monotonic rise in AUC as $k \to 0$ and the layer ordering replicate the $k \leq 50$ results, confirming the hockey-stick pattern is not an artifact of the trajectory length filter.}
  \label{fig:lookahead_max15}
\end{figure*}

\begin{figure*}[h]
  \centering
  \begin{subfigure}[b]{0.48\textwidth}
    \includegraphics[width=\linewidth]{figures/figures/laguna_xs2_pro_full_pooled_pooled_pro/currently_correct_lookahead_pro_max50.pdf}
  \end{subfigure}\hfill
  \begin{subfigure}[b]{0.48\textwidth}
    \includegraphics[width=\linewidth]{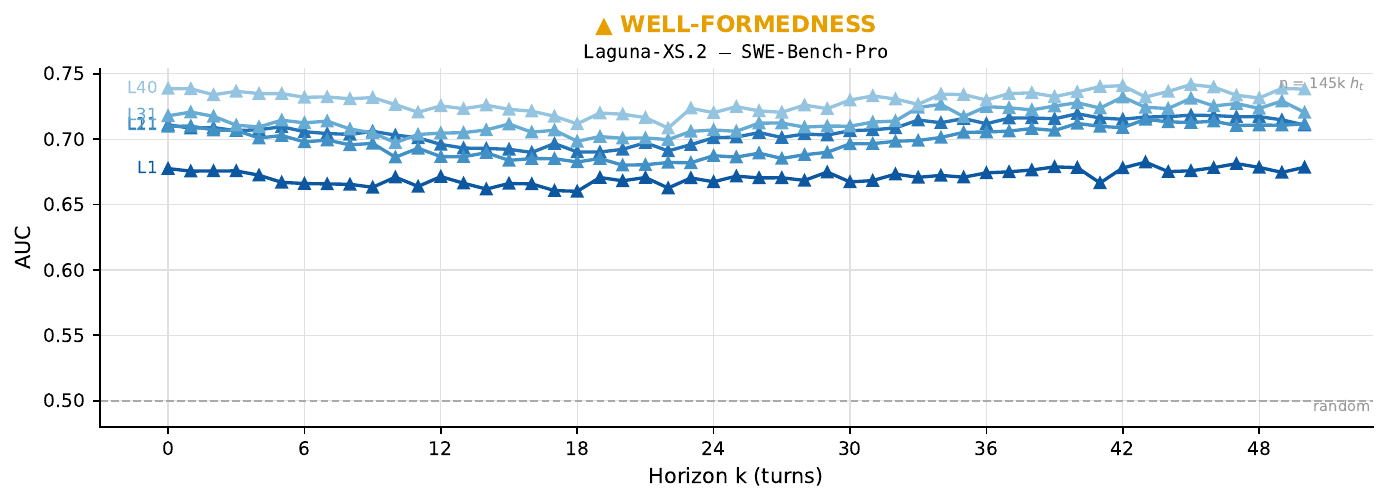}
  \end{subfigure}\\[0.5em]
  \begin{subfigure}[b]{0.48\textwidth}
    \includegraphics[width=\linewidth]{figures/figures/laguna_xs2_pro_full_pooled_pooled_pro/currently_reduces_failing_lookahead_pro_max50.pdf}
  \end{subfigure}\hfill
  \begin{subfigure}[b]{0.48\textwidth}
    \includegraphics[width=\linewidth]{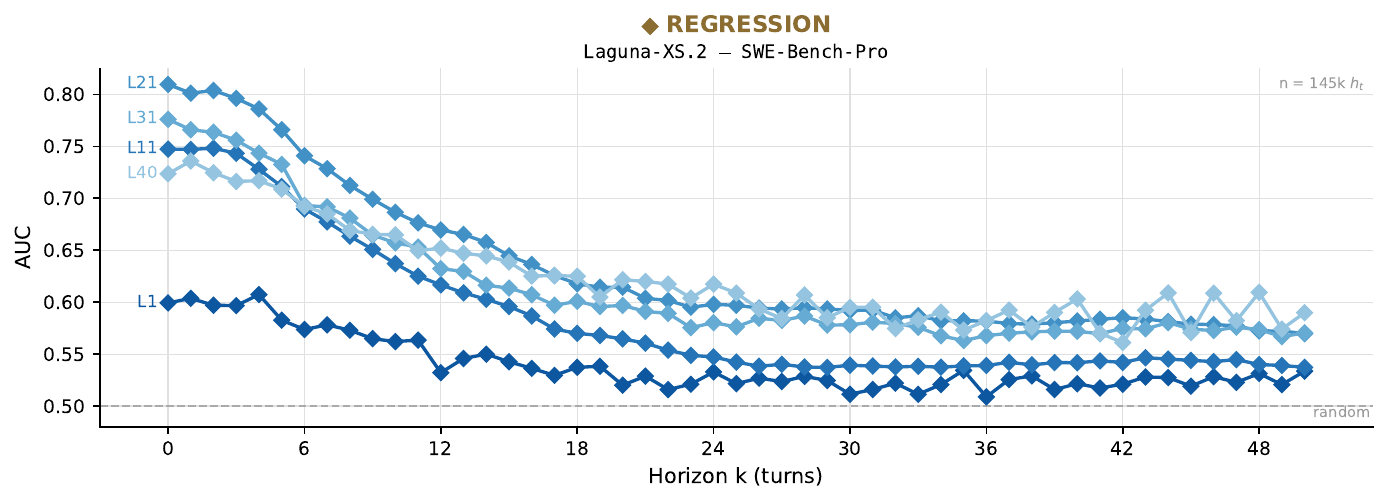}
  \end{subfigure}
  \caption{Lookahead AUC for all four probes on \texttt{Laguna-XS.2} (\texttt{SWE-Bench-Pro}), $k \leq 50$. Panel layout matches \Cref{fig:lookahead_verified}. The hockey-stick pattern replicates across this harder, multi-language benchmark.}
  \label{fig:lookahead_pro}
\end{figure*}

\begin{figure*}[h]
  \centering
  \begin{subfigure}[b]{0.48\textwidth}
    \includegraphics[width=\linewidth]{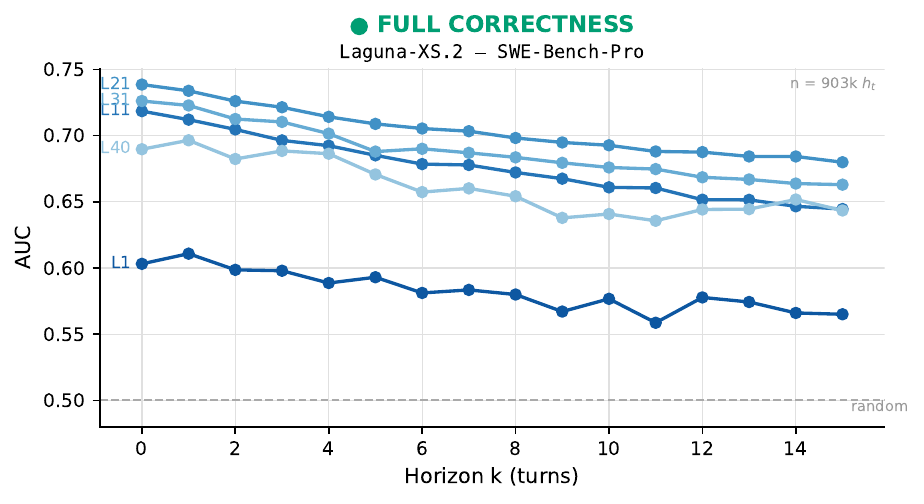}
  \end{subfigure}\hfill
  \begin{subfigure}[b]{0.48\textwidth}
    \includegraphics[width=\linewidth]{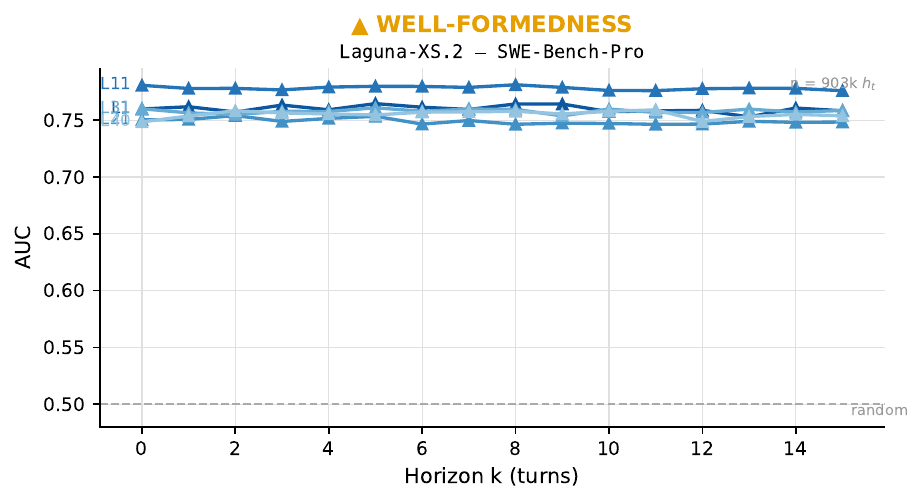}
  \end{subfigure}\\[0.5em]
  \begin{subfigure}[b]{0.48\textwidth}
    \includegraphics[width=\linewidth]{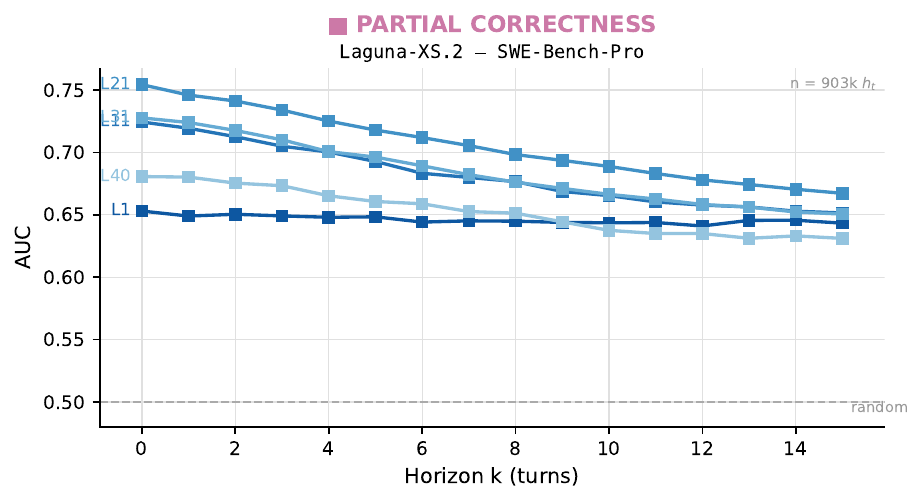}
  \end{subfigure}\hfill
  \begin{subfigure}[b]{0.48\textwidth}
    \includegraphics[width=\linewidth]{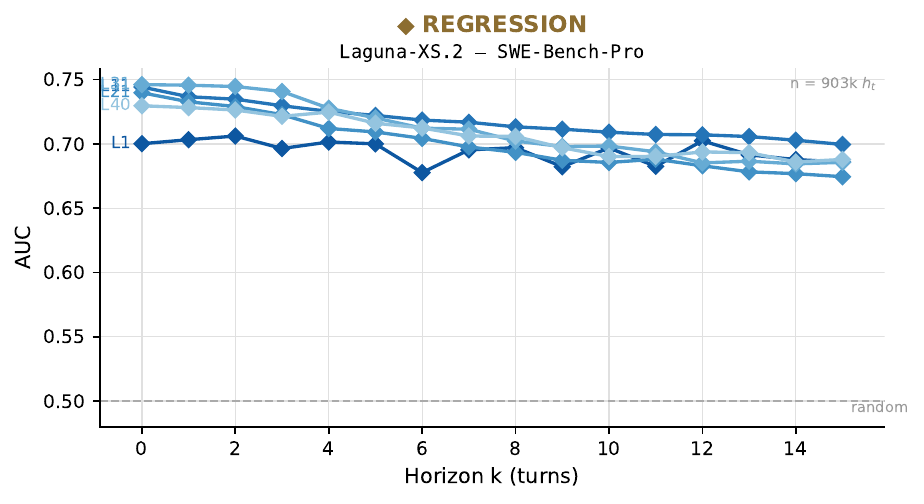}
  \end{subfigure}
  \caption{Lookahead AUC for all four probes on \texttt{Laguna-XS.2} (\texttt{SWE-Bench-Pro}), $k \leq 15$, using the full trajectory sample without the length filter. Panel layout matches \Cref{fig:lookahead_verified}. Results are consistent with the $k \leq 50$ analysis on this benchmark.}
  \label{fig:lookahead_pro_max15}
\end{figure*}

\begin{figure*}[h]
  \centering
  \begin{subfigure}[b]{0.48\textwidth}
    \includegraphics[width=\linewidth]{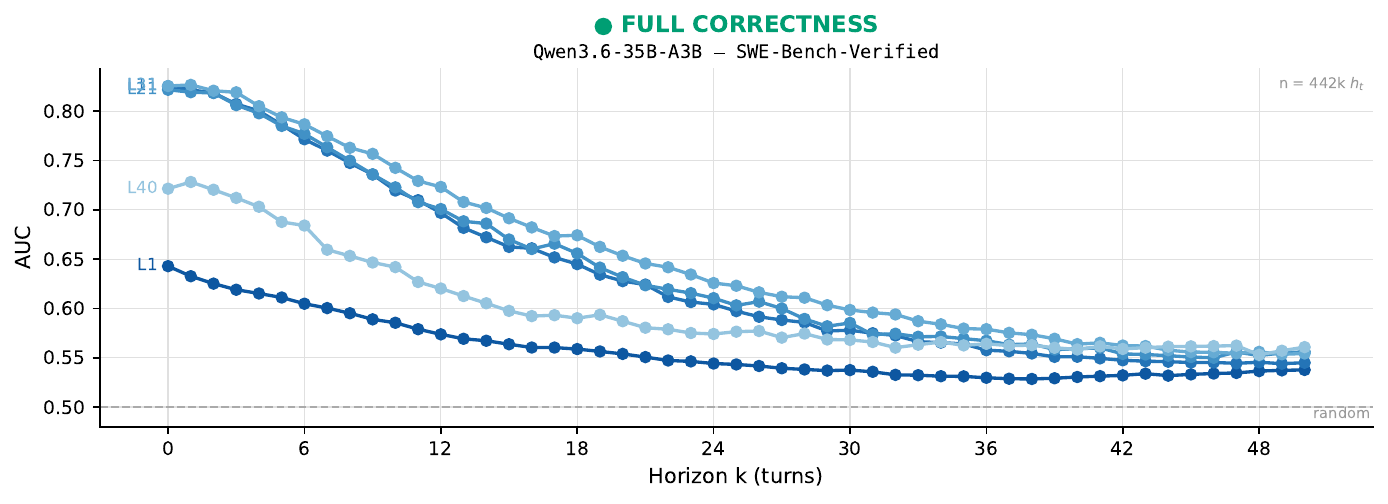}
  \end{subfigure}\hfill
  \begin{subfigure}[b]{0.48\textwidth}
    \includegraphics[width=\linewidth]{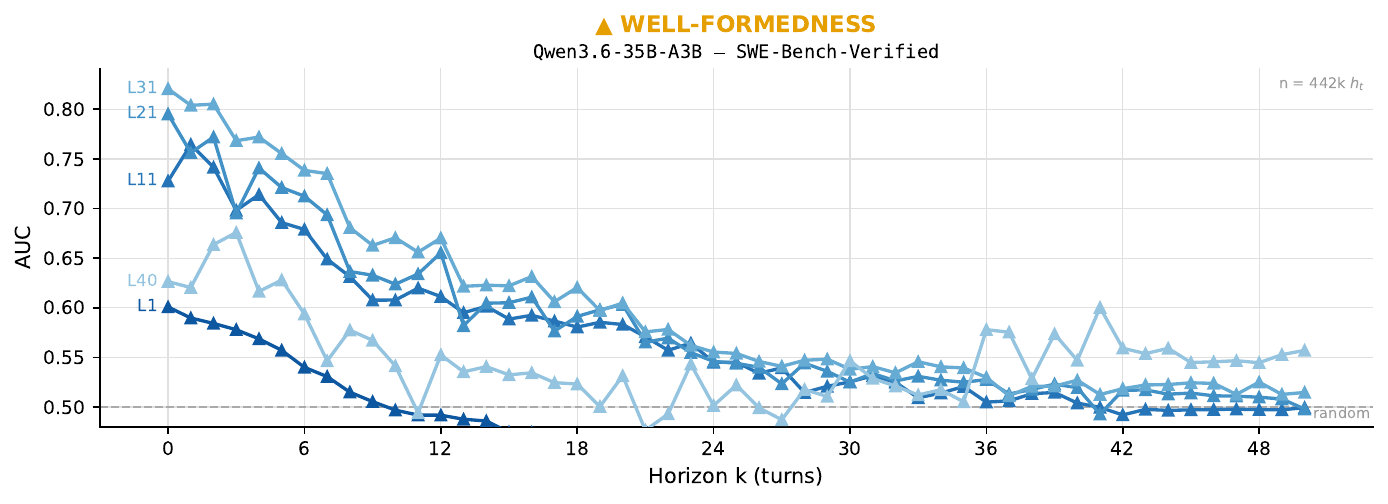}
  \end{subfigure}\\[0.5em]
  \begin{subfigure}[b]{0.48\textwidth}
    \includegraphics[width=\linewidth]{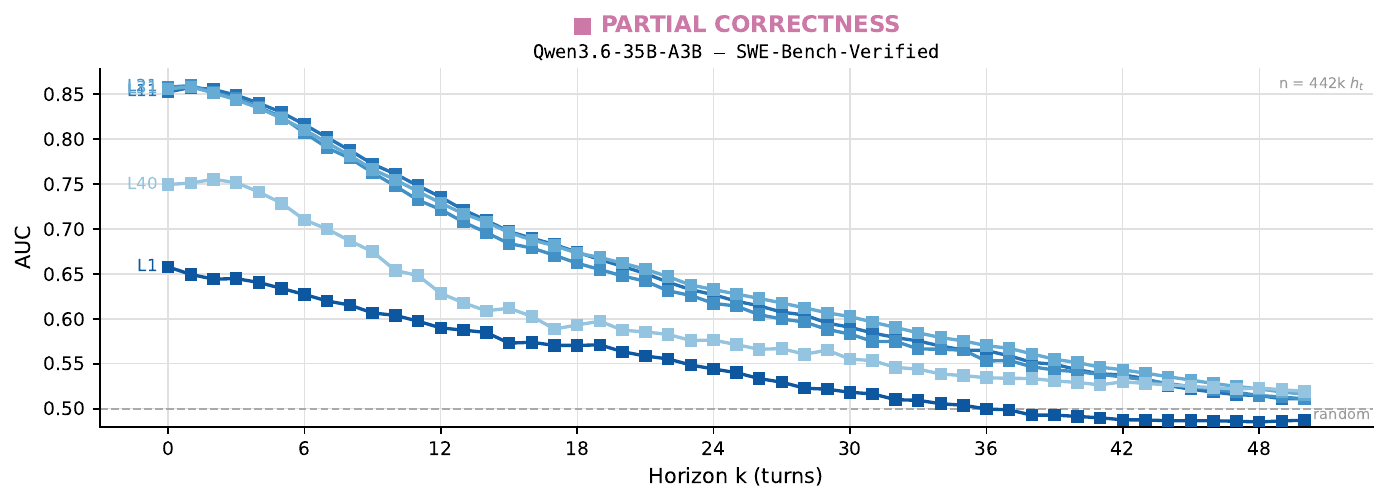}
  \end{subfigure}\hfill
  \begin{subfigure}[b]{0.48\textwidth}
    \includegraphics[width=\linewidth]{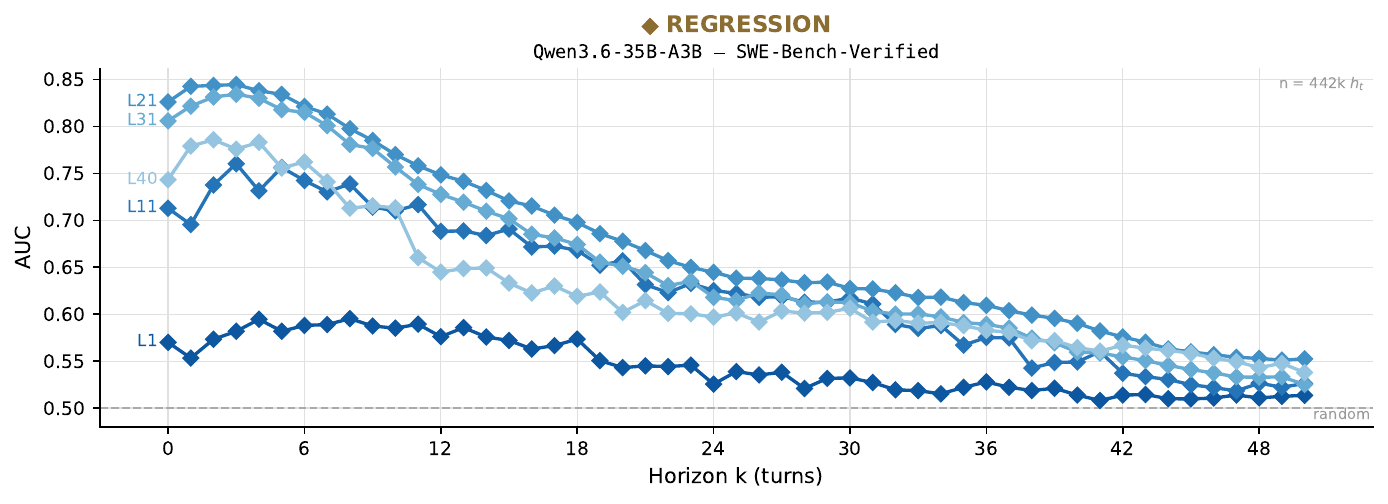}
  \end{subfigure}
  \caption{Lookahead AUC for all four probes on \texttt{Qwen3.6-35B-A3B} (\texttt{SWE-Bench-Verified}), $k \leq 50$. Panel layout matches \Cref{fig:lookahead_verified}. The hockey-stick pattern and layer ordering replicate at larger model scale.}
  \label{fig:lookahead_qwen_max50}
\end{figure*}

\begin{figure*}[h]
  \centering
  \begin{subfigure}[b]{0.48\textwidth}
    \includegraphics[width=\linewidth]{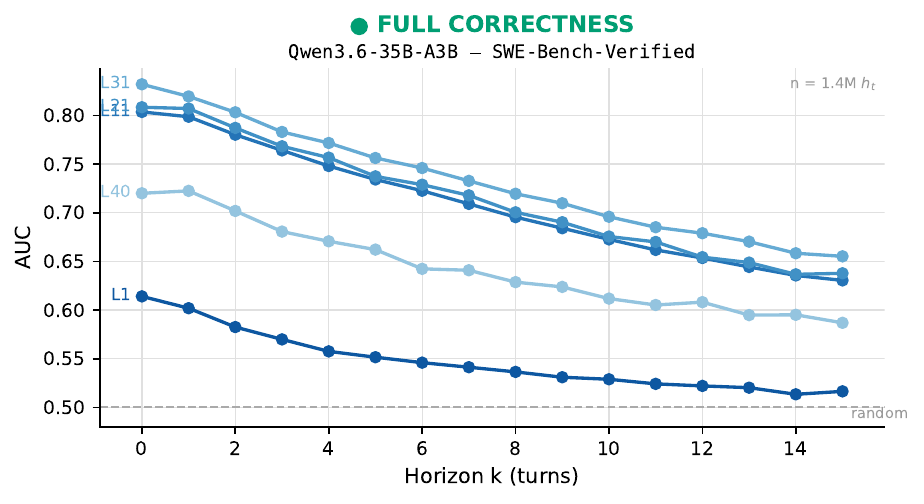}
  \end{subfigure}\hfill
  \begin{subfigure}[b]{0.48\textwidth}
    \includegraphics[width=\linewidth]{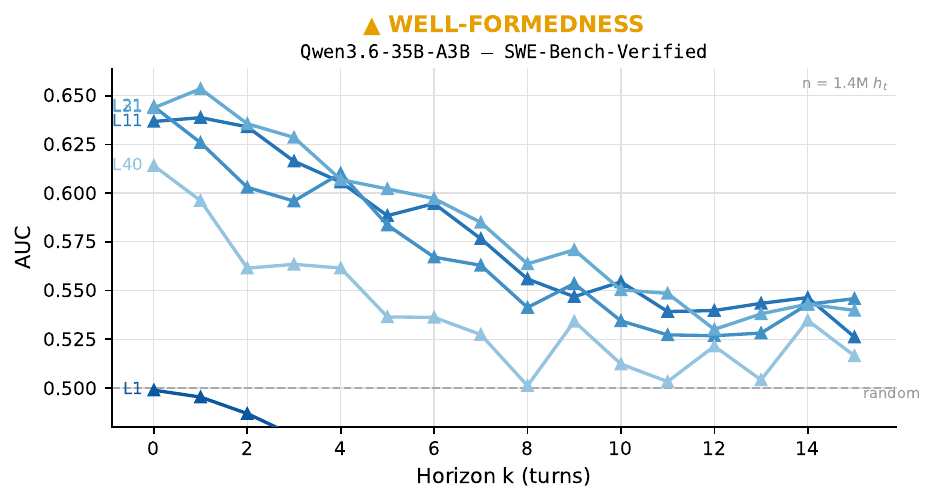}
  \end{subfigure}\\[0.5em]
  \begin{subfigure}[b]{0.48\textwidth}
    \includegraphics[width=\linewidth]{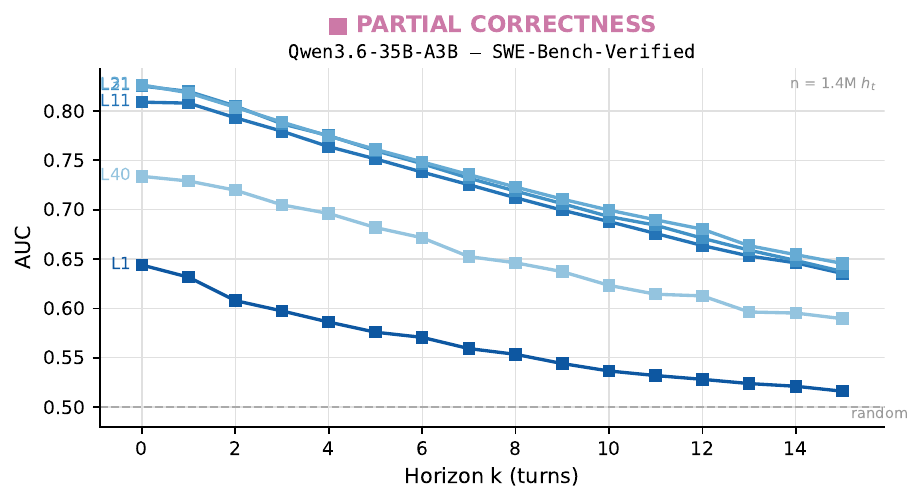}
  \end{subfigure}\hfill
  \begin{subfigure}[b]{0.48\textwidth}
    \includegraphics[width=\linewidth]{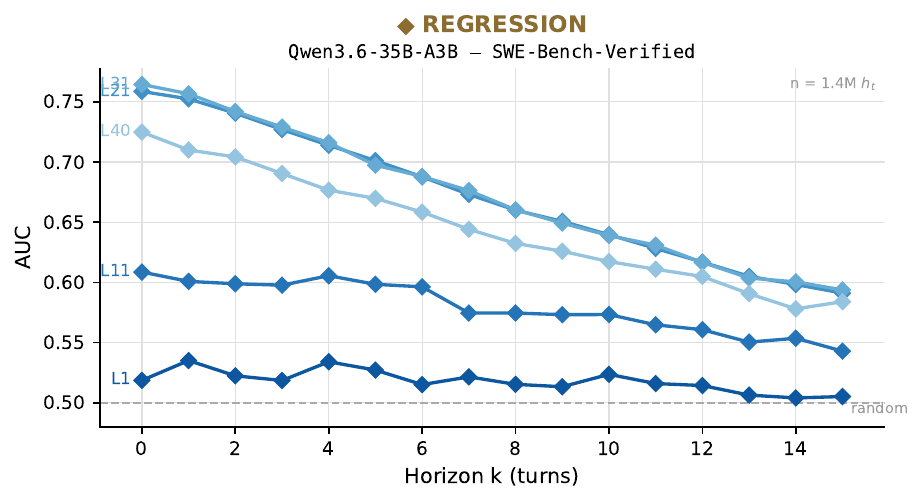}
  \end{subfigure}
  \caption{Lookahead AUC for all four probes on \texttt{Qwen3.6-35B-A3B} (\texttt{SWE-Bench-Verified}), $k \leq 15$, using the full trajectory sample without the length filter. Panel layout matches \Cref{fig:lookahead_verified}. Results are consistent with the $k \leq 50$ analysis for this model.}
  \label{fig:lookahead_qwen_max15}
\end{figure*}

\begin{figure*}[h]
  \centering
  \begin{subfigure}[b]{0.48\textwidth}
    \includegraphics[width=\linewidth]{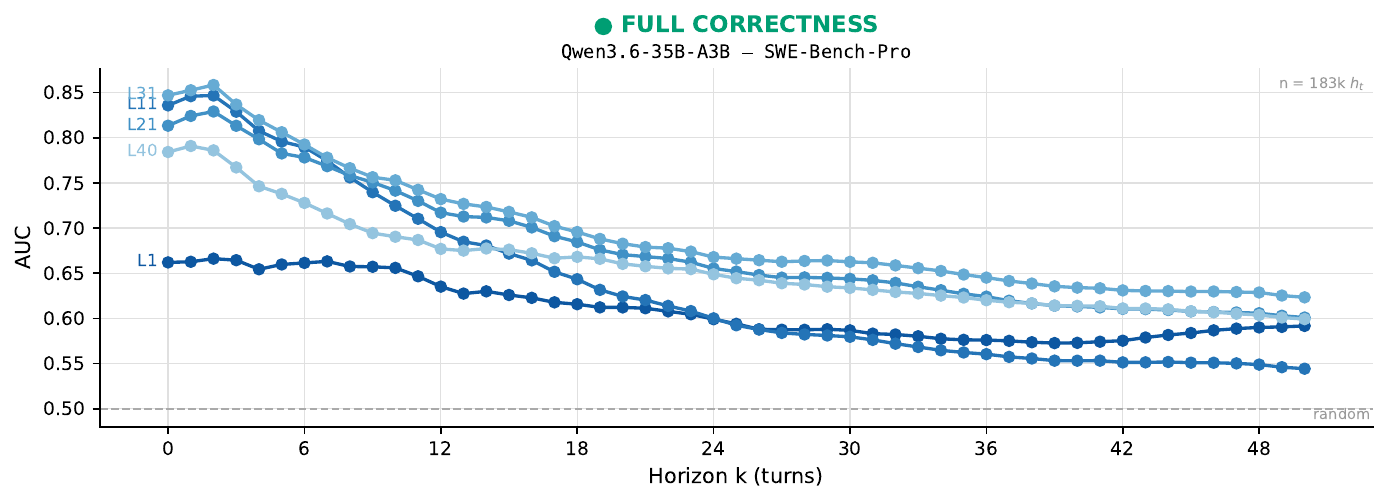}
  \end{subfigure}\hfill
  \begin{subfigure}[b]{0.48\textwidth}
    \includegraphics[width=\linewidth]{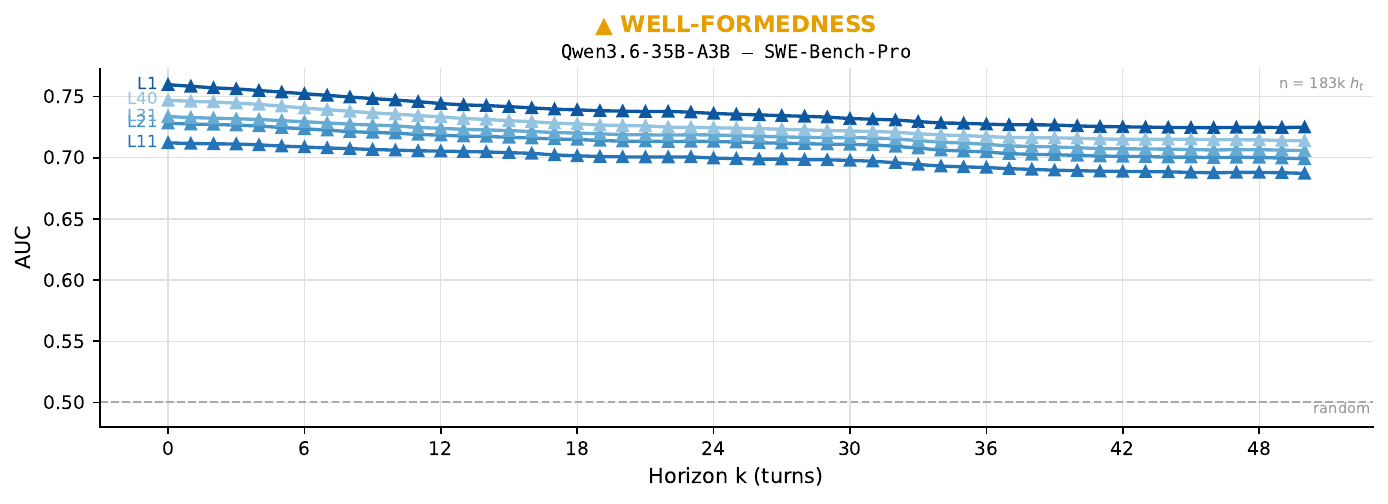}
  \end{subfigure}\\[0.5em]
  \begin{subfigure}[b]{0.48\textwidth}
    \includegraphics[width=\linewidth]{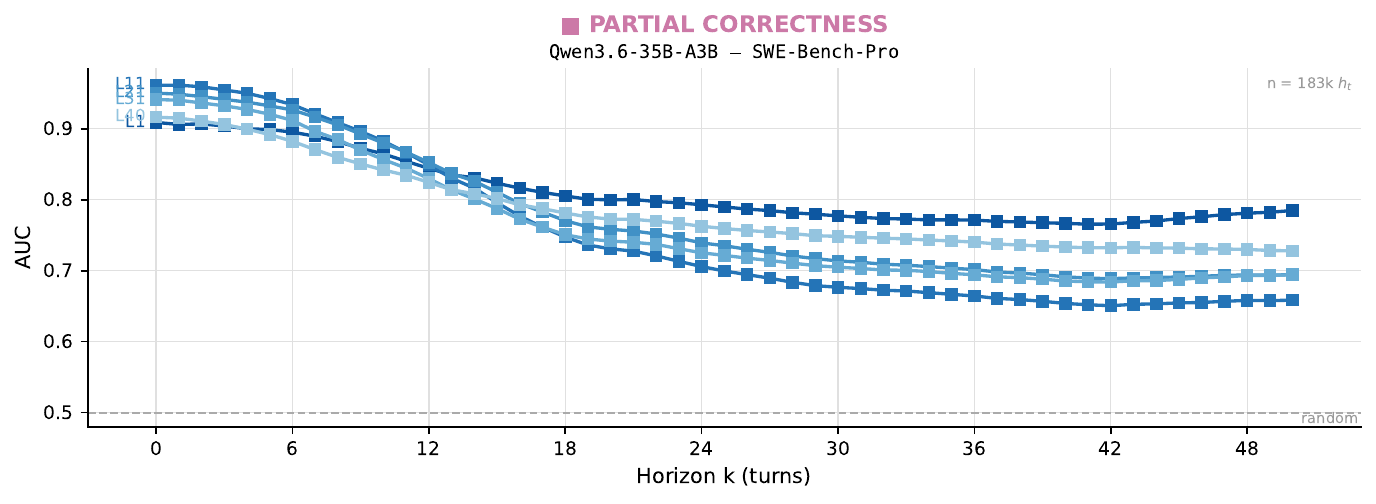}
  \end{subfigure}\hfill
  \begin{subfigure}[b]{0.48\textwidth}
    \includegraphics[width=\linewidth]{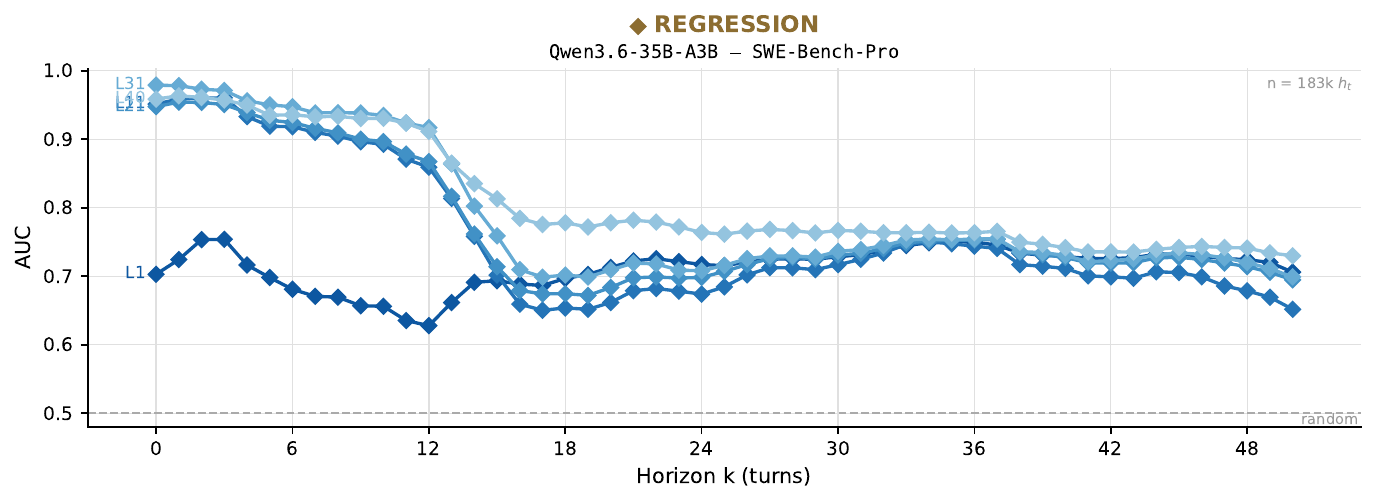}
  \end{subfigure}
  \caption{Lookahead AUC for all four probes on \texttt{Qwen3.6-35B-A3B} (\texttt{SWE-Bench-Pro}), $k \leq 50$. Panel layout matches \Cref{fig:lookahead_verified}. The hockey-stick pattern and layer ordering replicate at larger model scale on this harder, multi-language benchmark.}
  \label{fig:lookahead_qwen_pro_max50}
\end{figure*}

\begin{figure*}[h]
  \centering
  \begin{subfigure}[b]{0.48\textwidth}
    \includegraphics[width=\linewidth]{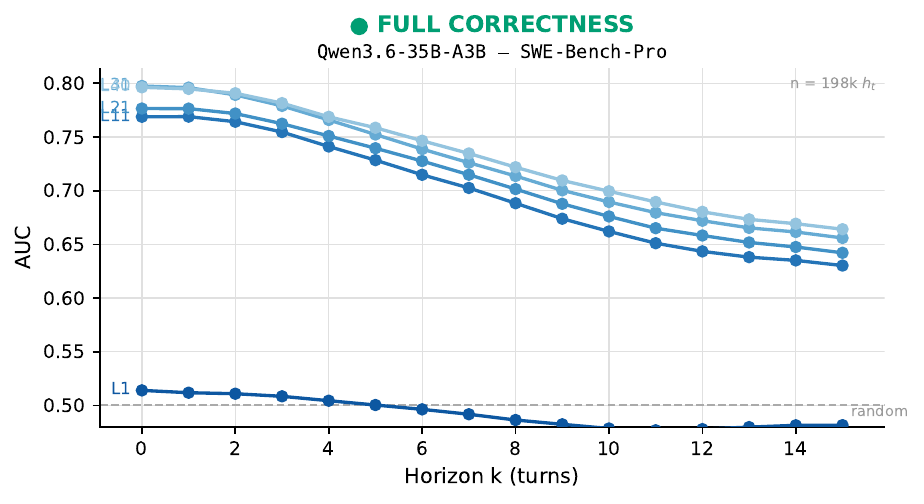}
  \end{subfigure}\hfill
  \begin{subfigure}[b]{0.48\textwidth}
    \includegraphics[width=\linewidth]{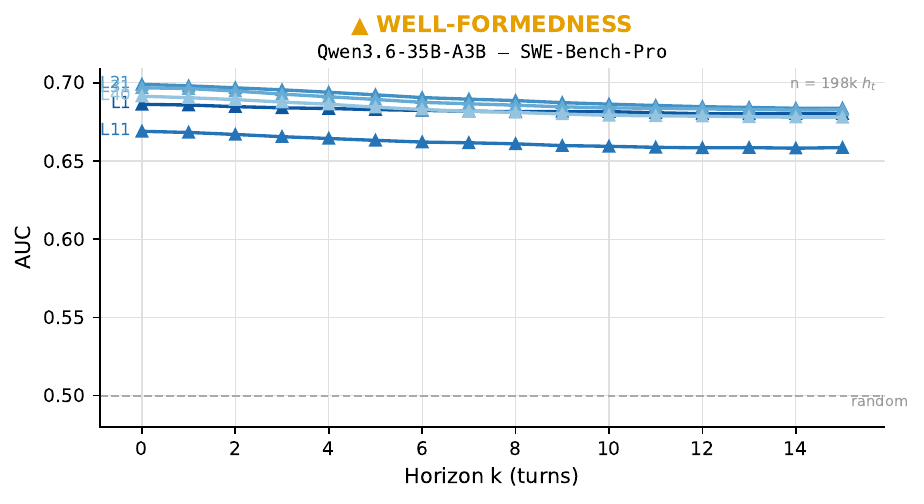}
  \end{subfigure}\\[0.5em]
  \begin{subfigure}[b]{0.48\textwidth}
    \includegraphics[width=\linewidth]{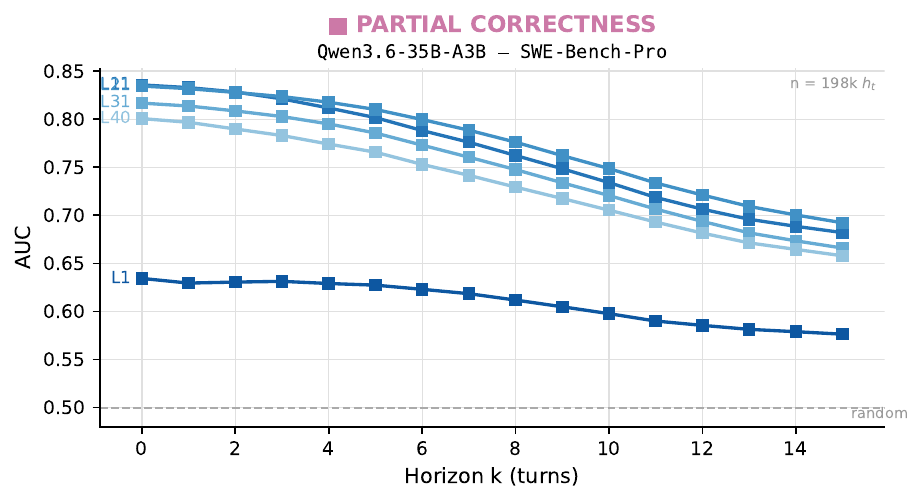}
  \end{subfigure}\hfill
  \begin{subfigure}[b]{0.48\textwidth}
    \includegraphics[width=\linewidth]{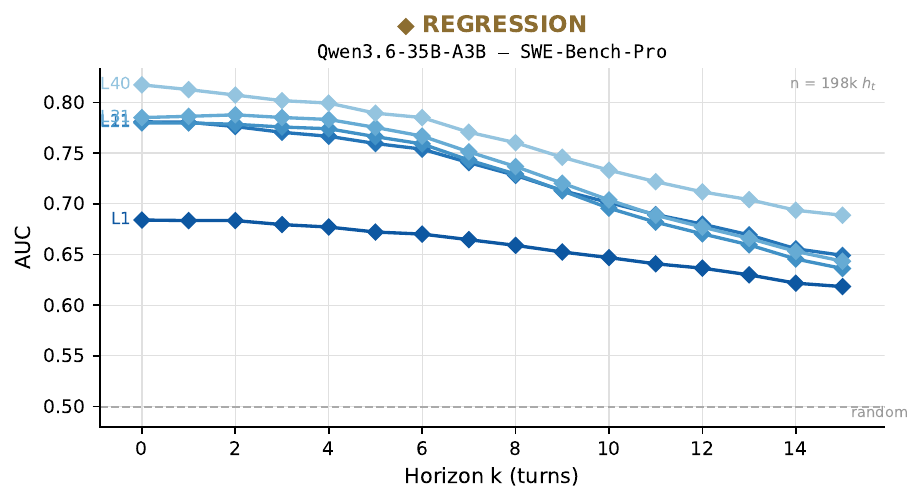}
  \end{subfigure}
  \caption{Lookahead AUC for all four probes on \texttt{Qwen3.6-35B-A3B} (\texttt{SWE-Bench-Pro}), $k \leq 15$, using the full trajectory sample without the length filter. Panel layout matches \Cref{fig:lookahead_verified}. Results are consistent with the $k \leq 50$ analysis on this benchmark.}
  \label{fig:lookahead_qwen_pro_max15}
\end{figure*}

\subsection{Example Latent Trajectories}
\label{app:example_trajectories}

\begin{figure}[t]
  \centering
  \begin{subfigure}[b]{0.48\textwidth}
    \includegraphics[width=\linewidth]{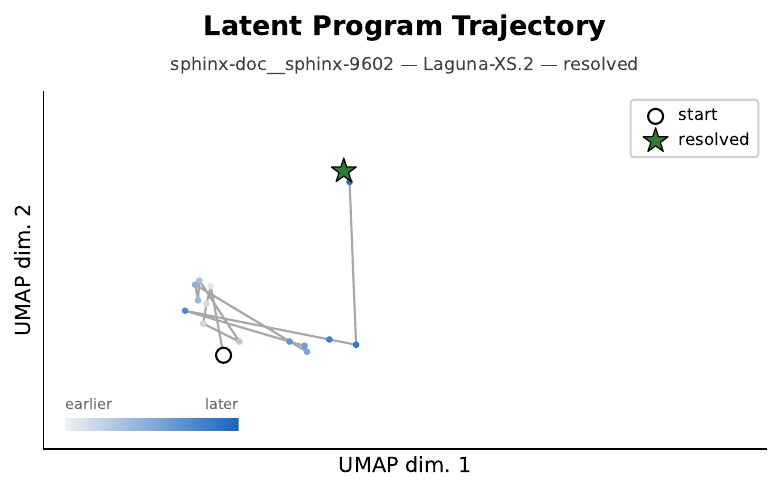}
  \end{subfigure}\hfill
  \begin{subfigure}[b]{0.48\textwidth}
    \includegraphics[width=\linewidth]{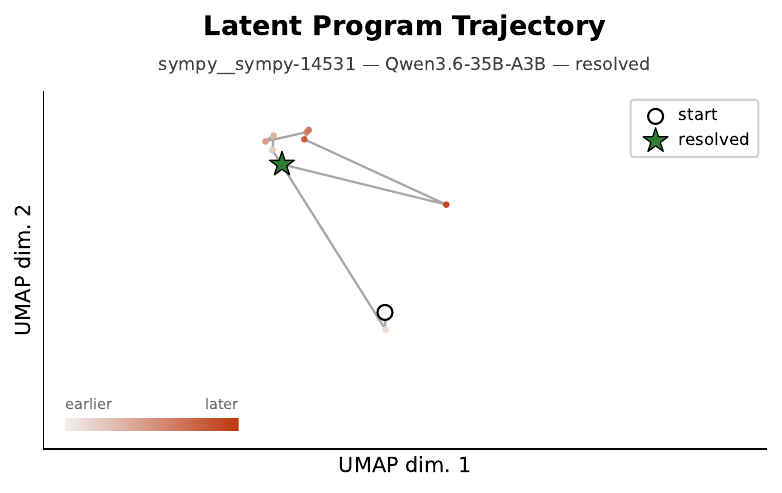}
  \end{subfigure}
  \caption{Two example resolved trajectories, hand-selected from a small set of long (many-edit), resolved rollouts for legibility, shown at layer 20 (best or near-best by probe validation AUC for both models). Left: \texttt{Laguna-XS.2} on \texttt{sphinx-doc/sphinx-9602}. Right: \texttt{Qwen3.6-35B-A3B} on \texttt{sympy/sympy-14531}. Each trajectory is reduced from its dense, stride-5 token-level activations to one point per code edit plus a start and an end marker, obtained by mean-pooling all captured hidden states within each corresponding span (before the first edit, between consecutive edits, and after the last edit). Color encodes step order (light = earlier, dark = later) in each model's palette color; the white circle marks the start and the star marks the final, resolved state. These examples are illustrative only and are not intended as quantitative evidence.}
  \label{fig:example_trajectories}
\end{figure}

\end{document}